\documentclass[letterpaper, 10 pt, journal, twoside]{IEEEtran} 

\IEEEoverridecommandlockouts                            %

\usepackage{amssymb}
\usepackage{amsmath}
\usepackage{mathtools}
\usepackage{algorithm}
\usepackage{algpseudocode}
\usepackage{tikz}
\usepackage{layouts}
\usepackage[font=footnotesize]{caption}
\usepackage{subcaption}
\usepackage{placeins}
\usepackage{todonotes}
\usetikzlibrary{shapes.geometric, arrows}
\usepackage{url}
\usepackage{lipsum}
\usepackage{cite}
\usepackage[hidelinks]{hyperref}

\usepackage{tensor}
\usepackage{stackengine}
\usepackage{etoolbox}
\usepackage{fnpos}
\usepackage{fancyhdr}
\usepackage{pifont}
\usepackage{setspace}
\usepackage{multicol}
\usepackage{multirow}
\usepackage{graphicx}
\usepackage{makecell}
\usepackage{float}
\usepackage{stfloats}
\usepackage{booktabs}

\makeatletter
\patchcmd{\ALG@step}{\addtocounter{ALG@line}{1}}{\refstepcounter{ALG@line}}{}{}
\newcommand{\ALG@lineautorefname}{Line}
\makeatother

\newcommand{\pnorm}[2]{\left\lVert#1\right\rVert_{#2}}

\newcommand{\Real}{\mathbb R}

\newcommand{\task}{\tau}
\newcommand{\q}{\mathbf{q}}
\renewcommand{\v}{\mathbf{v}}
\newcommand{\x}{\mathbf{x}}

\renewcommand{\u}{\mathbf{u}}
\newcommand{\X}{\mathcal{X}}
\newcommand{\U}{\mathcal{U}}

\newcommand{\A}{\mathbf{A}}
\newcommand{\B}{\mathbf{B}}

\newcommand{\pt}{\mathbf{P}}     %
\newcommand{\rot}{\mathbf{R}}     %

\newcommand{\SE}[1]{\text{SE}(#1)}

\newcommand{\SO}[1]{\text{SO}(#1)}
\newcommand{\so}[1]{\mathfrak{so}(#1)}
\newcommand{\Log}{\operatorname{Log}}

\newcommand{\defeq}{\vcentcolon=}
\newcommand{\Q}{\mathbf{Q}}

\DeclareMathAlphabet\mathbfcal{OMS}{cmsy}{b}{n}
\DeclareMathOperator*{\argmax}{arg\,max}
\DeclareMathOperator*{\argmin}{arg\,min}
\DeclareMathOperator*{\lexmin}{lex\,min}

\definecolor{LightCyan}{rgb}{0.88,1,1}
\definecolor{ashgrey}{rgb}{0.7, 0.75, 0.71}

\def\Vec#1{\!\!\hbox{$#1$\kern-0.38em\lower0.85em\hbox{$\vec{}\,$}}\,}%
\newcommand{\bbm}{\begin{bmatrix}}
	\newcommand{\ebm}{\end{bmatrix}}
\DeclareMathAlphabet{\mbf}{OT1}{ptm}{b}{n}

\usepackage{color, colortbl}

\newtheorem{preexample}{\sf\bfseries Example}

\newtheorem{prethm}{\sf\bfseries Theorem}

\newtheorem{prelem}{\sf\bfseries Lemma}

\newtheorem{preprop}{\sf\bfseries Proposition}

\author{Jingxing Qian$^{*,1}$, Xintong Du$^{*,1}$, Jun Yang$^{2}$, Siqi Zhou$^{3,4}$,
Steven L. Waslander$^{5}$, and Angela P. Schoellig$^{3}$%
\thanks{*Jingxing Qian and Xintong Du contributed equally.}%
\thanks{$^{1}$University of Toronto Institute for Aerospace Studies (UTIAS) and the Vector Institute for Artificial Intelligence, Canada.
Emails: \nolinkurl{jingxing.qian@robotics.utias.utoronto.ca}, \nolinkurl{xintong.du@robotics.utias.utoronto.ca}.}%
\thanks{$^{2}$Epson Canada Ltd., Toronto, ON L3R 6G3, Canada.
Email: \nolinkurl{Jun.Yang@ea.epson.com}.}%
\thanks{$^{3}$Learning Systems and Robotics Lab at the Technical University of Munich and the Munich Institute for Robotics and Machine Intelligence (MIRMI), Germany.
Email: \nolinkurl{angela.schoellig@tum.de}.}%
\thanks{$^{4}$School of Computing Science, Faculty of Applied Sciences, Simon Fraser University, Burnaby, BC, Canada.
Email: \nolinkurl{siqi@sfu.ca}.}%
\thanks{$^{5}$Institute for Aerospace Studies (UTIAS) and Robotics Institute, University of Toronto, Canada.
Email: \nolinkurl{steven.waslander@robotics.utias.utoronto.ca}.}%
}

\title{\LARGE \bf

Object- and Task-Aware Active Perception  for Safe Navigation of Mobile Manipulators in Semi-Static Environments
}

\begin{document}

\maketitle

\setcounter{topnumber}{1}
\renewcommand{\topfraction}{0.99}
\renewcommand{\textfraction}{0.01}

\setlength{\textfloatsep}{10pt} %

\begin{abstract}
Mobile manipulators must navigate safely in cluttered, partially unknown environments across logistics, manufacturing, and service applications. Semi-static changes between robot visits can lead to outdated scene knowledge and increased collision risk. We propose an active perception framework that enables a high degree-of-freedom (DoF) mobile manipulator to inspect uncertain or safety-critical regions with its hand camera during navigation. A probabilistic, object- and task-aware next-best-view (NBV) formulation prioritizes observations near the desired path and around uncertain or potentially changing objects. The perception objective is integrated into hierarchical-task model predictive control (HTMPC), exploiting kinematic redundancy for viewpoint adjustment while preserving the higher priority of base-trajectory tracking. The system unifies navigation, map maintenance, and active perception in a closed loop. Real-world tests show improved map consistency, reduced task-dependent map entropy, and lower navigation risk than passive or non-task-aware baselines. Additional simulation and real-world sequences demonstrate responses to controlled semi-static scene changes.
\end{abstract}

\section{Introduction}\label{sec:intro}

Mobile manipulators have gained increasing interest in a wide range of applications, including household assistance, logistics, and industrial automation. In these settings, robots are often required to execute navigation tasks across large workspaces over extended periods of time~\cite{thakar_manipulator_2022}. However, many deployment sites are cluttered, partially unknown, and semi-static, where objects may appear, disappear, or shift between robot visits. Clutter introduces frequent occlusions, blind corners, and narrow passages, while semi-static scene changes cause previously observed regions to become outdated. These challenges are particularly pronounced when the robot must perceive the environment through a limited field of view (FoV). In such cases, occlusions and blind spots can prevent changes from being detected until the robot is in close proximity, leaving new obstacles in previously traversable areas undetected. This can lead to delayed reactions, increased collision risk, and reduced navigation efficiency.

To operate safely and robustly in such cluttered, semi-static environments, a mobile manipulator must not only navigate toward its desired goal but also maintain an up-to-date, consistent understanding of the scene over time. However, this requirement is not well supported by most existing navigation pipelines, which typically assume a static world with an offline-constructed map for downstream planning and control~\cite{Cadena2016PastPA}, or only track short-term, highly dynamic objects between consecutive frames. Recent research on long-term and semi-static mapping has made important progress toward maintaining consistent environment representations across repeated traversals~\cite{panoptictsdf, QianChatrathPOCD}. By incorporating semantic and object-level reasoning, these approaches can detect scene inconsistencies and accommodate long-term changes, thereby improving map quality and robustness over time. Nevertheless, such methods remain fundamentally passive: they depend on observations collected along the robot’s nominal trajectory, and thus can only update the regions that the robot happens to observe. In cluttered environments, this is often insufficient, especially in occluded regions where up-to-date scene knowledge is most needed.

\begin{figure}[t!]
\centering
\includegraphics[width=0.99\linewidth]{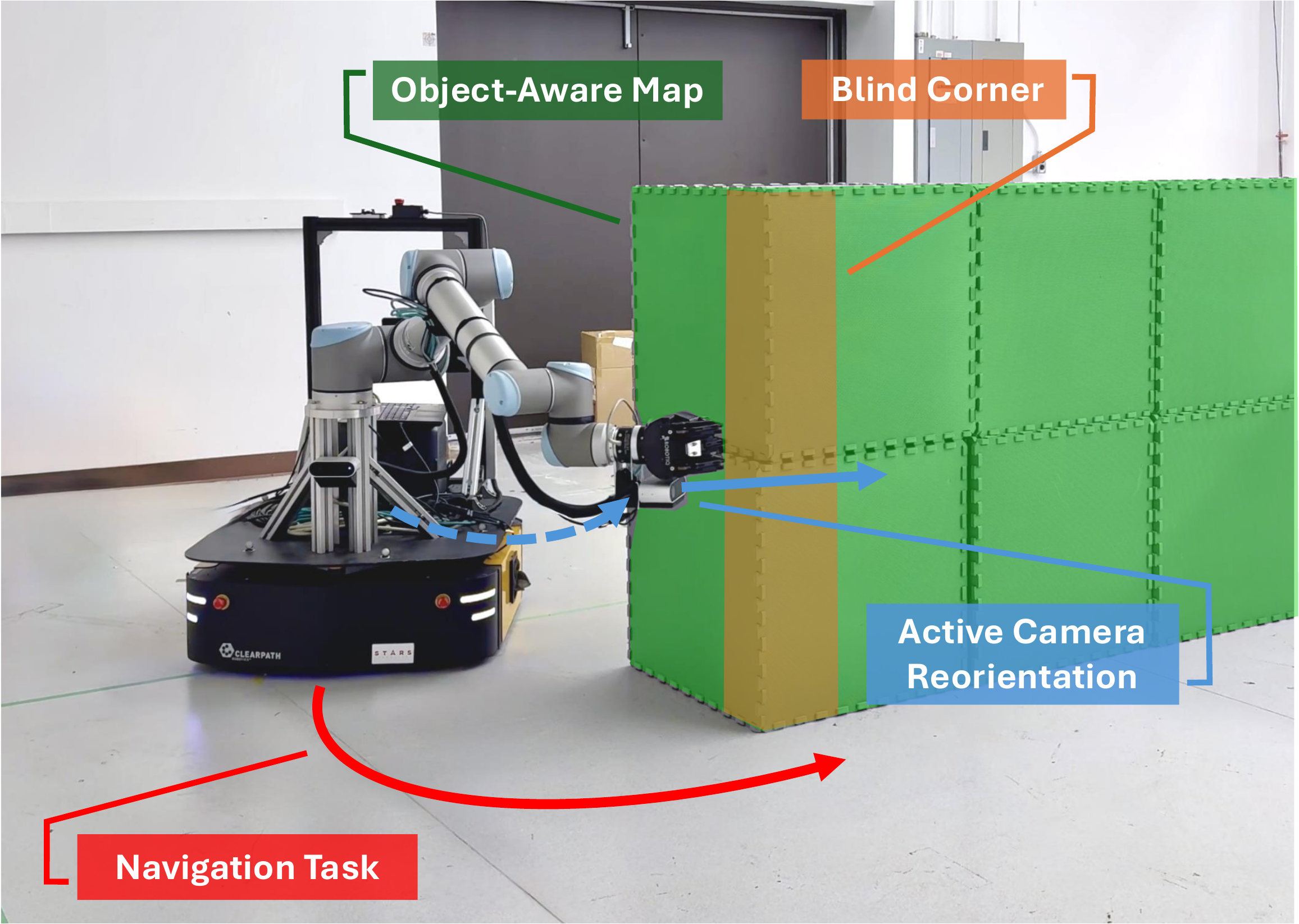}
  \caption{
A snapshot of the proposed active perception and navigation framework for high-degree-of-freedom (DoF) mobile manipulators in action. Equipped with a hand camera, the robot operates in cluttered, semi-static environments. It uses its kinematic redundancy to reposition the camera around obstacles and observe uncertain or occluded regions near its navigation path. See the multimedia attachment or \href{https://tinyurl.com/nbv-for-mm-nav}{online videos}\protect\footnotemark[6] for demonstrations. \vspace{-2mm}}
\label{fig:cover}
\end{figure}
\footnotetext[6]{\raggedright Video demonstrations: \url{https://tinyurl.com/nbv-for-mm-nav}.}
\setcounter{footnote}{6}

\begin{figure*}[!t]
\centering
\includegraphics[width=0.99\textwidth]{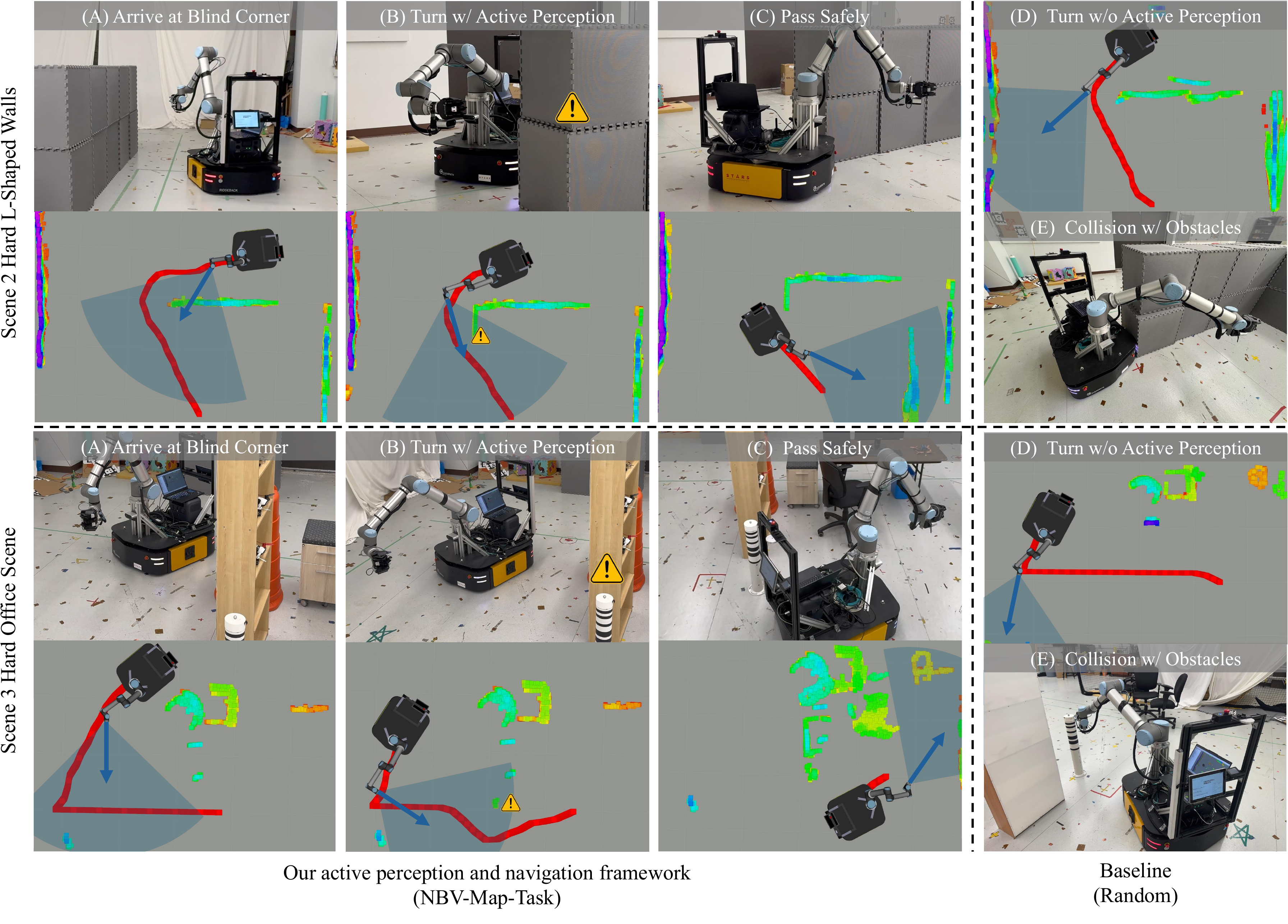}
  \caption{Qualitative demonstration of the proposed active perception and navigation framework for mobile manipulation in action (left) compared with a baseline method that randomly points the hand camera (right). Red curves represent the planned base paths. Shaded blue regions represent the hand camera’s FoV. Yellow exclamation marks highlight hidden hazards such as a blind corner (top) and a hidden pole (bottom). \vspace{-3mm}}
\label{fig:qual_against_baseline}
\end{figure*}

This limitation is especially critical because not all map regions affect navigation equally. Outdated knowledge in task-relevant and safety-critical regions, such as areas along the planned path, narrow passages, and blind corners, has a disproportionate impact on the robot’s ability to move safely and efficiently. As seen in Figure~\ref{fig:cover}, when the robot makes a sharp turn around the boxes, it is most important for the robot to observe the region around its navigation path behind the orange-shaded blind corner, which is not immediately observable by the fixed camera on the robot base. An outdated or incomplete map in these regions can lead to overly conservative behavior, unnecessary replanning, getting stuck, or collisions. Therefore, map maintenance is particularly valuable in regions where it improves the robot’s ability to make safe task decisions. In this work, we refer to this principle as \emph{task-awareness}: observations should be prioritized according to their relevance to the robot’s current navigation task, especially when they reduce uncertainty in map regions that affect safe progress toward the goal. %

Active perception provides a means to acquire these observations during task execution. Related approaches couple perception and motion planning for control and collision avoidance~\cite{Tordesillas2021PANTHERPT,Zheng2025ActivePC}. Next-best-view (NBV) planning has also been studied for exploration~\cite{Yamauchi1997AFA}, object seeking~\cite{Bogenberger2025WhereDI}, inspection~\cite{Naazare2022OnlineNP}, pose estimation~\cite{yang2025active}, and grasping~\cite{Ma2024ActivePF}. A camera mounted on the manipulator hand provides a natural mechanism for viewpoint control. However, many NBV methods target short-horizon operation in static scenes~\cite{di2024learning,hanlon2024active} or geometric completeness~\cite{lim2023fisher,chen2024gennbv}. They generally do not jointly consider long-term scene changes, object-level map consistency, and relevance to the current navigation task.

For semi-static environments, this omission is particularly important because scene changes often occur at the level of objects: furniture may be moved, shelves may be added, or obstacles may partially block previously traversable space. A partial mismatch at the visible edge of a cabinet can indicate that the cabinet has moved, making its previously mapped region worth checking again. As a result, identifying which objects or occupied structures are likely to have changed and which of them matter more for the robot’s navigation task is critical for effective map maintenance and safe task execution. This motivates an object-aware active perception strategy in which viewpoint selection is informed by uncertainty in the maintained semi-static map, especially when such uncertainty may affect path clearance. Task awareness then determines whether this potential change matters to the next maneuver.

Moreover, because the hand camera is mounted on a manipulator, viewpoint selection cannot be treated independently from navigation. The robot must realize informative camera motions without compromising progress on its primary navigation task. We therefore use hierarchical-task model predictive control (HTMPC) as the mechanism for enforcing this priority structure and realizing the selected viewpoints through the robot's kinematic redundancy~\cite{du_hierarchicaltaskmodel_2024a,duPerceptiveHierarchicalTaskMPC2026}.

In this work, we aim to enable safe mobile manipulation in cluttered, semi-static environments over extended periods of time through integrated active perception and whole-body control. Specifically, we develop an object- and task-aware active perception framework that combines the planned navigation path with uncertainty in the maintained semi-static map. Our approach enables a mobile manipulator to exploit its kinematic redundancy to direct a hand camera toward uncertain, occluded, and safety-critical regions while continuing to execute its primary navigation task. By actively updating the map where it matters most, the robot can better detect and adapt to incremental scene changes and improve map consistency in task-relevant areas. We deploy the system on a real mobile manipulator and evaluate reconstruction accuracy, task-dependent map entropy, and navigation safety against passive and task- or object-unaware baselines. Figure~\ref{fig:qual_against_baseline} illustrates the resulting behavior in cluttered scenes. Our main contributions are as follows:

\begin{itemize}
    \item We formulate a probabilistic next-best-view (NBV) objective that uses task-dependent map entropy and path-conditioned grid weights to prioritize uncertain or potentially changing regions along the planned path.
    \item We introduce a hierarchical-task control framework that assigns higher priority to navigation than to active perception within a two-stage optimization problem. This allows the robot to exploit its kinematic redundancy to gather informative observations while bounding the base-tracking cost during camera-task optimization.
    \item We demonstrate the system on a mobile manipulator with a hand camera in cluttered indoor environments. In the tested static scenes, the proposed system reduces task-dependent map entropy, improves map consistency, and lowers navigation risk compared with passive or non-task-aware baselines. Additional scene-change sequences demonstrate the closed-loop map-maintenance behavior.
\end{itemize}

\vspace{-2mm}
\section{Related Work}\label{sec:relatedwork}

\subsection{Semantic and Object-Aware Mapping}\label{sec:lit-obj-mapping}

Early mapping systems rely on sparse feature points~\cite{ORBSLAM3_TRO} or dense volumetric grids such as occupancy grids~\cite{octomap} and signed or Euclidean distance fields (SDFs/EDFs)~\cite{voxblox} to construct an accurate geometric map. However, purely geometric maps lack semantic context, limiting their expressiveness in complex environments. More recent work has incorporated semantic and object-level knowledge into geometric maps. Early approaches introduced objects into mapping by using predefined CAD models~\cite{slam++}. Subsequent object-aware methods further decompose the scene into semantic or object submaps on-the-fly~\cite{Rnz2018MaskFusionRR,grinvald2019volumetric,Rosinol20icra-Kimera}, allowing higher-level reasoning about object relationships and scene structure for downstream planning tasks~\cite{Gu2023ConceptGraphsO3,hughes2022hydra}.

Despite these advances, most semantic and object-aware mapping systems continue to prioritize geometric accuracy under a static-world assumption. When deployed over long time horizons in real-world semi-static environments, these methods can accumulate inconsistencies, leading to map corruption, degraded localization, and increased safety risks.

\subsection{Change Detection in Semi-Static Environments}\label{sec:lit-scene-changes}

To relax the static-world assumption, several strategies have been proposed to handle scene dynamics. A common approach identifies potentially dynamic objects based on prior semantic knowledge and treats them as outliers~\cite{DM-SLAM,Schmid2023DynabloxRD}. While efficient, this strategy can lead to information loss and instability when large parts of the scene have changed~\cite{tipaldi_lifelong_2013}. Another strategy jointly tracks the robot and objects~\cite{Rnz2018MaskFusionRR,TSDF++,Xu2019MIDFusionOO}, but typically relies on motion consistency across consecutive frames, making them less effective for semi-static environments, where changes occur between robot visits.

Detecting semi-static changes is critical for long-term autonomy, but has received less attention. Some recent approaches estimate object-level consistency based on overlap between observations and mapped objects~\cite{panoptictsdf,visinsMulti}. Other approaches leverage learned representations or scene graphs to predict object changes~\cite{Fu2023NeuSENS,Looper20223DVL}, use Bayesian formulations to propagate object consistency ~\cite{rosen2016towards,QianChatrathPOCD}, and jointly optimize robot state and object consistency~\cite{QianPOVSLAM}.

While these methods improve long-term map consistency, they are typically evaluated in open-loop mapping scenarios and rely on passive observations. As a result, changes occurring in occluded or task-critical regions may remain undetected until they lead to failure. An open question is how to couple probabilistic object-level change estimates with online viewpoint selection and control for safe navigation.

\subsection{Active Perception in Robotics}\label{sec:lit-active-perception}
Active perception deliberately selects sensor viewpoints to acquire observations that improve a robotic task~\cite{chen2011active}. For exploration and reconstruction, NBV planners commonly seek views that reveal unmapped space or reduce map uncertainty~\cite{Bircher2016RecedingNBV,isler2016information}. Naazare et al.~\cite{Naazare2022OnlineNP} extend this principle to exploration and inspection with an arm-mounted RGB-D camera, combining information gain with coverage of user-specified regions of interest. These approaches establish the utility of movable sensors, while the choice of which regions to inspect depends on the task being performed.

For navigation, Fisher-information methods select observations that improve localization~\cite{zhang2018perception,lim2023fisher}. The Fisher Information Field~\cite{zhang2019beyond,zhang2020fisher} represents the information available for visual localization and supports evaluation of candidate poses during motion planning. Learning-based methods instead predict informative viewpoints from data for localization~\cite{hanlon2024active,di2024learning} or exploration~\cite{chen2024gennbv}. Our objective uses map uncertainty and path relevance to choose observations of obstacles and free space that affect navigation.

Visibility-aware and risk-averse planners also account for sensing limitations and path-relevant uncertainty when planning motion at the cost of navigation efficiency~\cite{Kim2025VisibilityRRT,Khass2025RiskAverseNBV}. Our focus is maintaining object-level consistency across repeated visits and realizing informative hand-camera views under a navigation-first task hierarchy.

\subsection{Perception-based Mobile Manipulation}
\label{sec:lit-mm-in-unstructured-env}
Perception-based control frameworks, in which onboard sensor data continuously inform motion planning and control, have enabled mobile manipulators to operate in changing environments for tasks such as reactive grasping and obstacle avoidance~\cite{burgess-limerickArchitectureReactiveMobile2023, pankertPerceptiveModelPredictive2020, spahnCoupledMobileManipulation2021}.
In all these works, however, the coupling is one-directional: the robot perceives to inform control, but sensor viewpoints are determined passively by the task-driven motion rather than actively optimized.
Active perception has been explored in mobile manipulation for specific subtasks such as object recognition~\cite{xieCoupledActivePerception2024c} and grasp pose detection~\cite{jauhriActivePerceptiveMotionGeneration2024}. SPIN~\cite{Uppal2024SPIN} further learns coordinated perception, interaction, and navigation from onboard vision without constructing an environment map. Our setting emphasizes persistent object-map maintenance: observations acquired during navigation must reveal changes to a scene mapped on earlier visits, and the resulting map must inform subsequent motion.

Hierarchical task control offers a principled way to integrate active perception into whole-body motion planning for kinematically redundant robots. By establishing a strict order among tasks, the hierarchical task control framework leverages the robot's redundancy such that the lower-priority tasks are solved only in the null space of higher-priority ones~\cite{khatib_unified_1987, escande_hierarchical_2014}, avoiding conflicting motions inherent in the tasks.
HTMPC instantiates this idea in a model predictive control setting through nonlinear lexicographic optimization, which enables strictly ordered sequential mobile manipulation tasks~\cite{du_hierarchicaltaskmodel_2024a, duPerceptiveHierarchicalTaskMPC2026}.
In this work, we employ HTMPC with active perception as an explicit task in the hierarchy, secondary to the navigation task, resolving the inherent perception-control objective conflicts and creating a truly bidirectional perception-control loop: the robot plans its action based on the latest perceived map, and simultaneously optimizes its sensor placement to actively perceive the surrounding environment.
This enables safe whole-body navigation in unstructured, semi-static environments.

\section{Preliminaries}\label{sec:background}

\subsection{Object-Aware Semi-Static Mapping}\label{sec:bg-semistatic}
We consider semi-static environments in which most objects remain stationary over short time horizons, but may appear, disappear, or change location between repeated robot visits. In such settings, maintaining a globally consistent map requires not only estimating scene geometry from observations, but also reasoning about whether previously mapped objects are still valid. To achieve this, we build on the probabilistic object-level change detection framework in POCD~\cite{QianChatrathPOCD}, which represents the scene as an object library $\mathcal{O}=\{\mathbf{O}_i\}_{i=1\cdots I}$ and explicitly models the temporal and geometric consistency of each object. Each mapped object $\mathbf{O}_i$ forms an object-level submap containing:
\begin{itemize}
    \item a 3D position and a planar yaw angle in the global frame,
    \item an EDF voxel-grid submap $\mathcal{M}_i$ in the global frame,
    \item a state distribution $p(l_i,v_i)$, where $l_i\in\mathbb{R}$ models the object's geometric change and $v_i\in[0,1]$ models the object's consistency, with larger values indicating a higher probability that the mapped object remains valid.
\end{itemize}
This object-centric representation allows the system to reason about scene changes at the object level, which is more robust than treating map cells independently under partial observations and repeated traversals.

Following~\cite{QianChatrathPOCD}, the state distribution of object $\mathbf{O}_i$ at time step $k$ is parameterized as the product of a Gaussian distribution for the continuous geometric-change variable $l_i$ and a Beta distribution for the bounded consistency variable $v_i$:
\begin{equation}
\label{eq:bg-state-param}
\begin{aligned}
    p(l_i,v_i)
    &\defeq q(l_i,v_i \mid \mu_{i,k}, \sigma_{i,k}, \alpha_{i,k}, \beta_{i,k}) \\
    &\defeq \mathcal{N}(l_i \mid \mu_{i,k}, \sigma_{i,k}^2)
    \mathrm{Beta}(v_i \mid \alpha_{i,k}, \beta_{i,k}).
\end{aligned}
\end{equation}
Here, $\mu_{i,k}$ and $\sigma_{i,k}^2$ represent the current estimate and variance of the geometric change, while $\alpha_{i,k}$ and $\beta_{i,k}$ parameterize the consistency belief. The expectation
\begin{equation}
\label{eq:bg-consistency-est}
    \mathbb{E}[v_i] = \frac{\alpha_{i,k}}{\alpha_{i,k}+\beta_{i,k}}
\end{equation}
serves as the \emph{object consistency estimate}. It has been used in prior works for map maintenance and for downstream planning and control to indicate potentially changed objects around which the robot should behave cautiously~\cite{QianMPCCBF,Bogenberger2025WhereDI}. In this work, we additionally use the variance
\begin{equation}
\label{eq:bg-consistency-var}
    \mathrm{Var}[v_i]
    =
    \frac{\alpha_{i,k}\beta_{i,k}}
    {(\alpha_{i,k}+\beta_{i,k})^2(\alpha_{i,k}+\beta_{i,k}+1)}
\end{equation}
as an uncertainty measure for the object's consistency. Thus, $\mathbb{E}[v_i]$ captures the expected validity of the mapped object, while $\mathrm{Var}[v_i]$ captures uncertainty in that validity estimate.

For each matched object-observation pair, the measurement consists of two parts. First, the geometric consistency $\Delta_{i,k}$ is computed as the difference between the object and observation submaps. This term is modeled as an inlier when the object remains consistent with its current estimate, and as an outlier when the object has changed. The corresponding likelihood is written as the Gaussian-Uniform mixture~\cite{vogiatzis2011video,QianChatrathPOCD}
\begin{equation}
\label{eq:bg-geom-like}
\begin{aligned}
    p(\Delta_{i,k} \mid l_i, v_i)
    &\defeq
    v_i \mathcal{N}(\Delta_{i,k} \mid l_i, \tau^2) \\
    &\quad + (1-v_i)
    \mathcal{U}(\Delta_{i,k} \mid -\Delta_{\max}, \Delta_{\max}),
\end{aligned}
\end{equation}
where $\tau^2$ is the inlier measurement variance and $\Delta_{\max}$ bounds the outlier range. Second, a binary semantic label $s_{i,k}\in\{0,1\}$ provides a prior on whether the observed object is likely static or likely changing. This term enters the update as a weighted Bernoulli likelihood factor:
\begin{equation}
\label{eq:bg-stat-like}
    p(s_{i,k} \mid v_i)
    \defeq
    \mathrm{Bernoulli}(s_{i,k} \mid v_i)^{\kappa_{i,k}}.
\end{equation}
Here, $\kappa_{i,k}\geq0$ controls the relative weight of semantic evidence. The Bayesian update combines the new geometric and semantic evidence with the previous state estimate. Writing the prior at time step $k$ as $q(l_i,v_i \mid \mu_{i,k},\sigma_{i,k},\alpha_{i,k},\beta_{i,k})$, the posterior at time step $k+1$ satisfies
\begin{equation}
\label{eq:bg-bayes}
\begin{aligned}
    &q(l_i,v_i \mid \mu_{i,k+1},\sigma_{i,k+1},\alpha_{i,k+1},\beta_{i,k+1}) \\
    &\qquad \propto
    p(\Delta_{i,k} \mid l_i,v_i)\,
    p(s_{i,k} \mid v_i) \\
    &\qquad\quad \times
    q(l_i,v_i \mid \mu_{i,k},\sigma_{i,k},\alpha_{i,k},\beta_{i,k}).
\end{aligned}
\end{equation}
This posterior is obtained by multiplying the prior with the geometric-consistency and semantic-consistency likelihoods. It can then be approximated in closed form by matching moments with a single Gaussian-Beta distribution of the form in \eqref{eq:bg-state-param}~\cite{QianChatrathPOCD}. Operationally, matched observations that agree with the current object model typically increase the consistency estimate and reduce its uncertainty, while inconsistent or missing observations penalize consistency and increase uncertainty. If $\mathbb{E}[v_i]$ falls below a threshold, the object is treated as changed and reconstructed in the maintained map.

\subsection{Fisher Information and Differential Entropy for NBV}\label{sec:bg-nbv}
Next-best-view (NBV) planning selects sensing actions that are expected to reduce uncertainty relevant to a downstream task. A common way to formalize this idea is through information-theoretic metrics that quantify how informative a candidate observation is with respect to an unknown quantity of interest. Among these metrics, Fisher information provides a local measure of observation sensitivity~\cite{kay1993fundamentals}.

Consider a generic observation model
\begin{equation}
    Y \mid x \sim p(\cdot \mid x),
\end{equation}
where $Y\in\mathbb{R}$ is an observable random variable and $x\in\mathbb{R}^n$ is the latent quantity of interest. This generic notation can represent different active-perception quantities, such as a robot pose, map state, or scene parameter. In \autoref{sec:method-entropy}, we instantiate this template with the local robot base position used in the EDF likelihood. The Fisher score is defined as
\begin{equation}
    V = \nabla_x \log p(Y \mid x),
\end{equation}
and the Fisher Information Matrix (FIM) is defined as
\begin{equation}
\begin{aligned}
    \mathcal{F}(x)
    &= \mathbb{E}[VV^T] \\
    &= \mathbb{E}\big[(\nabla_x \log p(Y \mid x))
    (\nabla_x \log p(Y \mid x))^T\big].
\end{aligned}
\end{equation}
Under standard regularity conditions, the score has zero mean, so this expression coincides with the covariance of $V$. The FIM quantifies how sensitive the observation model is to changes in the latent quantity of interest, and is therefore a measure of local informativeness. However, NBV also requires comparing how much uncertainty remains under the current belief. For this reason, differential entropy is commonly used as a scalar measure of estimation uncertainty~\cite{cover2006elements}. Under a local Gaussian approximation, the covariance of the estimate of $x$
can be approximated by $\mathcal{F}(x)^{-1}$. The corresponding differential entropy is
\begin{equation}
    h_e(\mathcal{F}(x))
    =
    \frac{1}{2}\log\Big[(2\pi e)^n \det(\mathcal{F}(x)^{-1})\Big].
\end{equation}
Thus, under the local Gaussian approximation, uncertainty reduction can be measured by the decrease in the entropy induced by $\mathcal{F}(x)^{-1}$.

In active perception, information gain measures the reduction from prior to posterior uncertainty, while conventional expected information gain averages this reduction over future observations~\cite{isler2016information,zhang2020fisher}. Our approximate EIG measures the expected reduction in task-dependent map entropy over normalized grid-cell and rollout-time weights, using two representative observation outcomes per candidate view. The full construction is given in \autoref{sec:method-map}--\autoref{sec:method-eig}. This objective ranks viewpoints by their predicted information gain. Realizing the selected view also requires coordinating camera and base motion, which motivates the task hierarchy introduced next.

\subsection{Hierarchical-Task Model Predictive Control}\label{sec:bg-htmpc}
The HTMPC framework provides a principled way to coordinate multiple control objectives with strict task priorities over a finite prediction horizon. This is particularly useful for high-DoF mobile manipulators, where secondary objectives such as camera reorientation for active perception should exploit the robot's kinematic redundancy without degrading the primary navigation objective. For example, inspecting a blind corner may require the hand camera to move around an obstacle while the base follows its path. The controller must coordinate these motions according to their task priorities.

Given an ordered set of tasks $[\task_1,\task_2,\cdots,\task_L]$ in decreasing order of priority, HTMPC is formulated at each control time step $t_k$ as a lexicographic optimal control problem over a finite prediction time interval $\mathbb{T}_k\coloneqq[t_k, t_k+T_{\mathrm{mpc}}]$,
\begin{subequations}\label{eq:bg-htmpc}
\begin{align}
      \lexmin_{\x(\cdot),\u(\cdot)}\;& [\mathcal{J}_1,\mathcal{J}_2,\cdots,\mathcal{J}_L] \\
      \text{s.t.}\;& \dot{\x}(t)= \mathbf{f}_{\scriptstyle mdl}(\x(t), \u(t)), &&\forall t \in \mathbb{T}_k, \label{eq:bg-htmpc-dynamics}\\
      & \mathbf{g}(\x(t),\u(t))\leqslant \mathbf{0}, &&\forall t \in \mathbb{T}_k, \label{eq:bg-htmpc-nonlinear} \\
      & \x(t) \in \mathcal{X},\; \u(t) \in \mathcal{U}, &&\forall t \in \mathbb{T}_k, \label{eq:bg-htmpc-xu-cst}\\
      & \x(t_k)=\x_k, \label{eq:bg-htmpc-init-cst}
\end{align}
\end{subequations}
where $t$ is the continuous time variable, $\x(\cdot):\Real \xrightarrow{}\Real^{n_x}$ and $\u(\cdot):\Real \xrightarrow{}\Real^{n_u}$ are continuous state and input trajectories, and $\x_k=\x(t_k)$ is the current state. Each $\mathcal{J}_l$ is the cost associated with task $\task_l$, and the list of cost functions is minimized in lexicographic order through $\lexmin$, subject to robot dynamics $\mathbf{f}_{\scriptstyle mdl}$ and state- and input-dependent constraints $\mathbf{g}$, such as safety constraints. $\mathcal{X}$ and $\mathcal{U}$ are the feasible state and input sets. The optimization problem is solved at each control time step in a receding-horizon fashion.

Rather than combining all tasks into a single weighted objective, lexicographic optimization preserves strict priority among tasks. In practice, this is commonly implemented by solving a sequence of single-task MPC (STMPC) problems from the highest-priority task to the lowest-priority task. At iteration $l$, the problem can be written as
\begin{subequations}\label{eq:bg-stmpc}
\begin{align}
      \x_k^{l^*}(\cdot),\u_k^{l^*}(\cdot)
      = &\argmin_{\x(\cdot),\u(\cdot)}\; \mathcal{J}_l + \mathcal{J}_{\mathit{aux}} \\
      \text{s.t.}\;& \eqref{eq:bg-htmpc-dynamics}-\eqref{eq:bg-htmpc-init-cst}, \\
      & \mathcal{J}_i(\x(\cdot),\u(\cdot))\leqslant
      \mathcal{J}_i(\x_k^{i^*}(\cdot),\u_k^{i^*}(\cdot)), \\
      & \hspace{1.2em}\forall i=1,\cdots,l-1 \notag
\end{align}
\end{subequations}
where $(\x_k^{l^*}(\cdot),\u_k^{l^*}(\cdot))$ is the state-input trajectory obtained for task $\task_l$ over the prediction time interval $\mathbb{T}_k$, and $\mathcal{J}_{\mathit{aux}}$ is an auxiliary cost used to improve control performance and numerical conditioning. The task-cost constraints in \eqref{eq:bg-stmpc} prevent later optimizations from increasing the costs attained for previously solved higher-priority tasks. Consequently, only the remaining redundancy after satisfying the higher-priority task-cost bounds is used to realize lower-priority behaviors. HTMPC has been used to enforce the temporal order in sequential mobile manipulation tasks~\cite{du_hierarchicaltaskmodel_2024a, duPerceptiveHierarchicalTaskMPC2026}. In this paper, we use HTMPC to track the desired base motion as the primary task while using residual redundancy to realize perception-oriented hand camera motions. In the following sections, we drop the time dependence in state and control inputs wherever it is clear.

\section{System Description} \label{sec:system}

\begin{figure*}[t!]
\centering
\includegraphics[width=0.96\linewidth]{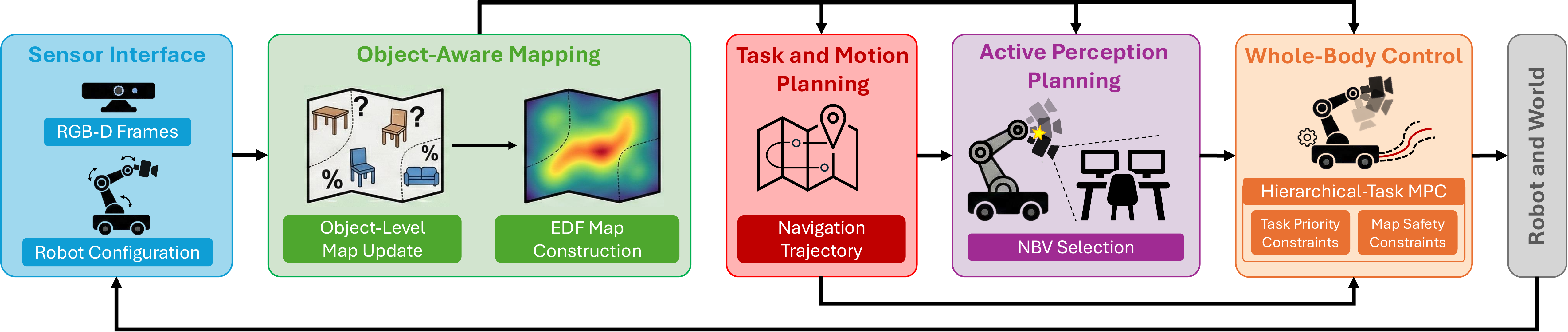}
  \caption{An overview of the proposed active perception and navigation framework for high-DoF mobile manipulators.\vspace{-5mm}}
\label{fig:system_overview}
\end{figure*}

\subsection{System Overview}\label{sec:system-overview}
\begin{figure}[b!]
\centering
\includegraphics[width=0.97\linewidth]{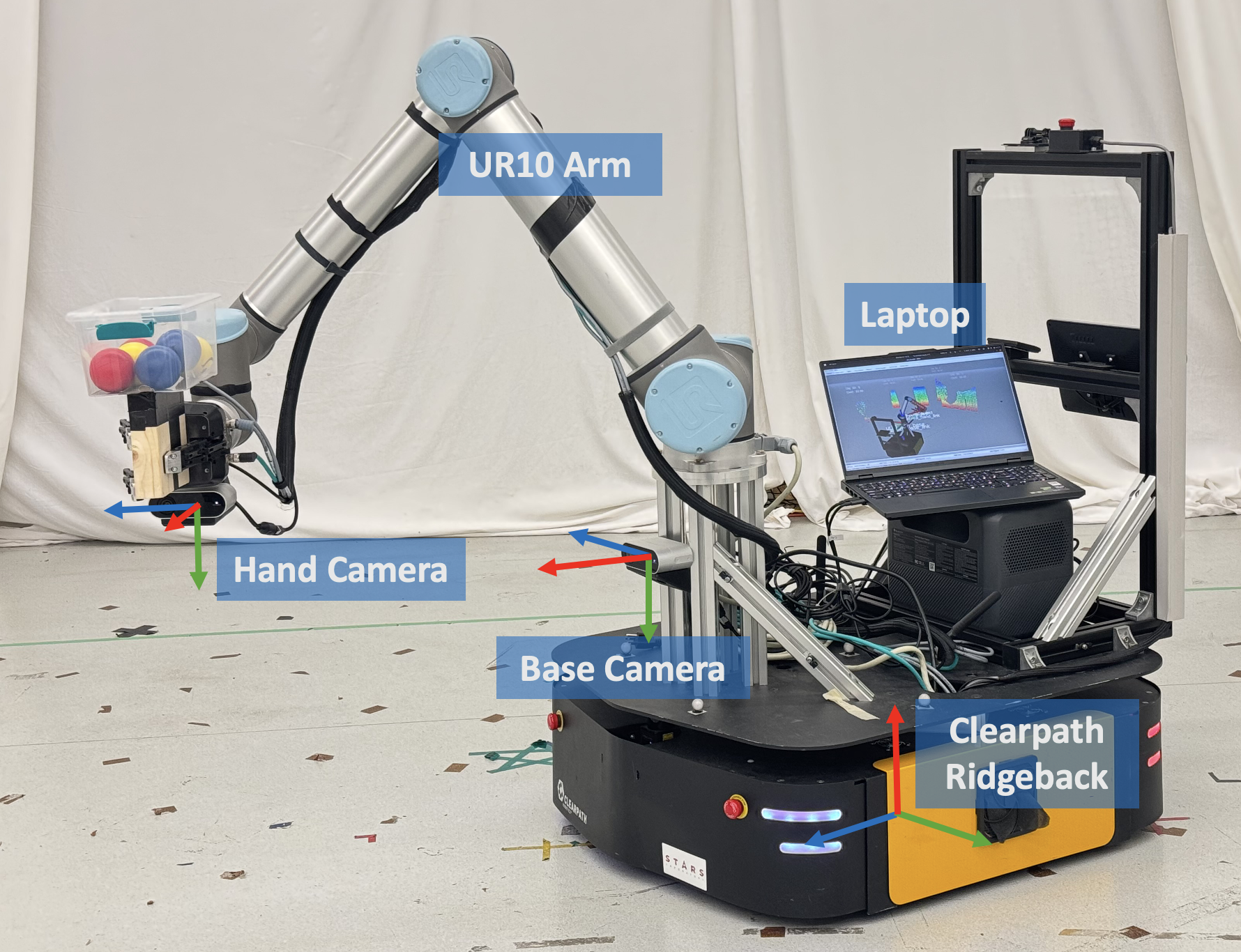}
  \caption{Our mobile manipulator setup. The system consists of a Universal Robots UR10 arm mounted on a Clearpath Robotics Ridgeback omnidirectional base. Two synchronized Orbbec Femto Bolt RGB-D cameras are installed, with one mounted on the base and the other mounted on the hand. An onboard laptop runs all the perception, planning and control algorithms. }
  \label{fig:robot_setup}
\end{figure}

We present an active perception and navigation framework for mobile manipulators operating in cluttered, unknown, and semi-static environments. Our robot platform, shown in \autoref{fig:robot_setup}, uses two synchronized RGB-D cameras: a \emph{base camera} rigidly mounted on the mobile base and a \emph{hand camera} mounted on the manipulator end effector. The robot is assigned a sparse set of navigation waypoints as its primary task, while the hand camera is allowed to reorient online to inspect uncertain, occluded, or safety-critical regions. The complete closed-loop pipeline is illustrated in \autoref{fig:system_overview}.

The system contains five modules: a sensor interface, an object-aware perception and mapping module, a path-planning module, an object- and task-aware NBV planner, and a planning and control module. At each time step $t_k$, camera, odometry, and joint measurements are synchronized to estimate the robot state, the map is updated, and a desired base path is planned from the current map and given waypoints. The active perception module then evaluates candidate viewpoints using their predicted reduction in task-dependent map entropy. The control module tracks the base path while realizing the selected hand camera objective, producing new observations that close the perception--action loop.

\subsection{Robot State and State Estimation}\label{sec:system-perception}
We first define the robot state used throughout the framework. Let $\q\in\Real^n$ denote the generalized coordinate, $\v\in\Real^m$ the generalized velocity, and $\u=\dot{\v}\in\Real^m$ the acceleration input. The full state is $\x = [\q^T,\v^T]^T$, with feasible state and input sets $\X$ and $\U$ respectively. The state is partitioned into mobile-base and arm components using subscripts $b$ and $a$. For our platform, $\mathbf{q}_{b,k},\mathbf{v}_{b,k}\in\Real^3$ describe the Ridgeback planar base pose and velocity, while $\mathbf{q}_{a,k},\mathbf{v}_{a,k}\in\Real^6$ describe the UR10 joint positions and velocities, resulting in $n=m=9$. At discrete time $t_k$, the robot state is written as
\begin{equation}
    \mathbf{x}_k =
    \begin{bmatrix}
        \mathbf{q}_{b,k}^T &
        \mathbf{q}_{a,k}^T &
        \mathbf{v}_{b,k}^T &
        \mathbf{v}_{a,k}^T
	    \end{bmatrix}^T .
\end{equation}

The sensor interface synchronizes camera, odometry, and joint measurements to produce an estimated robot state $\hat{\mathbf{x}}_k$. At time $t_k$, let $\mathbf{F}^{\textrm{base}}_k$ and $\mathbf{F}^{\textrm{hand}}_k$ denote the RGB-D frames captured by the base and hand cameras, $\mathbf{y}^{\textrm{odo}}_k$ the wheel odometry measurement, and $\mathbf{q}_{a,k},\mathbf{v}_{a,k}$ the arm encoder measurements. Together with the known camera extrinsics, these measurements form the estimator input.

The base pose and velocity $\mathbf{q}_{b,k}$ and $\mathbf{v}_{b,k}$ are estimated using a modified ORB-SLAM3~\cite{ORBSLAM3_TRO} with the base camera and wheel odometry. Wheel odometry improves robustness in feature-sparse regions and interpolates or extrapolates the base state to compensate for RGB-D frame-rate and latency limits. In parallel, the current hand camera pose is estimated using the measured arm joint states $\mathbf{q}_{a,k}$ and $\mathbf{v}_{a,k}$ through forward kinematics. The resulting state estimate $\hat{\mathbf{x}}_k$, together with the hand camera pose, is passed to the mapping, NBV, and control modules as their common robot state input. For the rest of the paper, we omit the hat and let $\mathbf{x}_k$ denote the estimated robot state for readability.

\subsection{Object-Aware Perception and Map Maintenance}\label{sec:system-mapping}
We expect the robot to operate in semi-static environments where rigid objects may appear, disappear, or shift over time. When deployed in a new environment, the robot starts with an empty map and incrementally updates it as it traverses the scene. To maintain an up-to-date scene representation, the RGB-D frames from both cameras are unprojected into the global frame, segmented into point-cloud observations $\mathcal{Y}_k=\{\mathbf{Y}_{k,p}\}_{p=1\cdots P}$ via geometric clustering, and merged into the mapped object library $\mathcal{O}$ through optimal association.

This module forms the map-maintenance core of the framework. The object representation and probabilistic state update follow the semi-static mapping model introduced in~\autoref{sec:bg-semistatic}. New observations spawn new objects, while matched observations update existing objects and propagate their state distributions through Bayesian inference. If an object's consistency expectation in \eqref{eq:bg-consistency-est} falls below a threshold, it is treated as changed and reconstructed in the maintained map. In addition to maintaining object-level consistency, this module produces the uncertainty estimates used in the downstream NBV module. Near mapped objects, the map uncertainty is inherited from the corresponding object consistency variance in \eqref{eq:bg-consistency-var}. In non-object regions, it is represented by a Beta occupancy uncertainty model. If an object is not observed for too long, its state model is additionally penalized by synthetic outlier evidence so that outdated objects become increasingly uncertain and are encouraged to be re-observed. The construction of these uncertainty fields is detailed in \autoref{sec:method-map}.

\vspace{-2mm}

\subsection{Object- and Task-Aware Next-Best-View Planning}\label{sec:system-nbv}
\begin{figure*}[t!]
\centering
\includegraphics[width=0.75\linewidth]{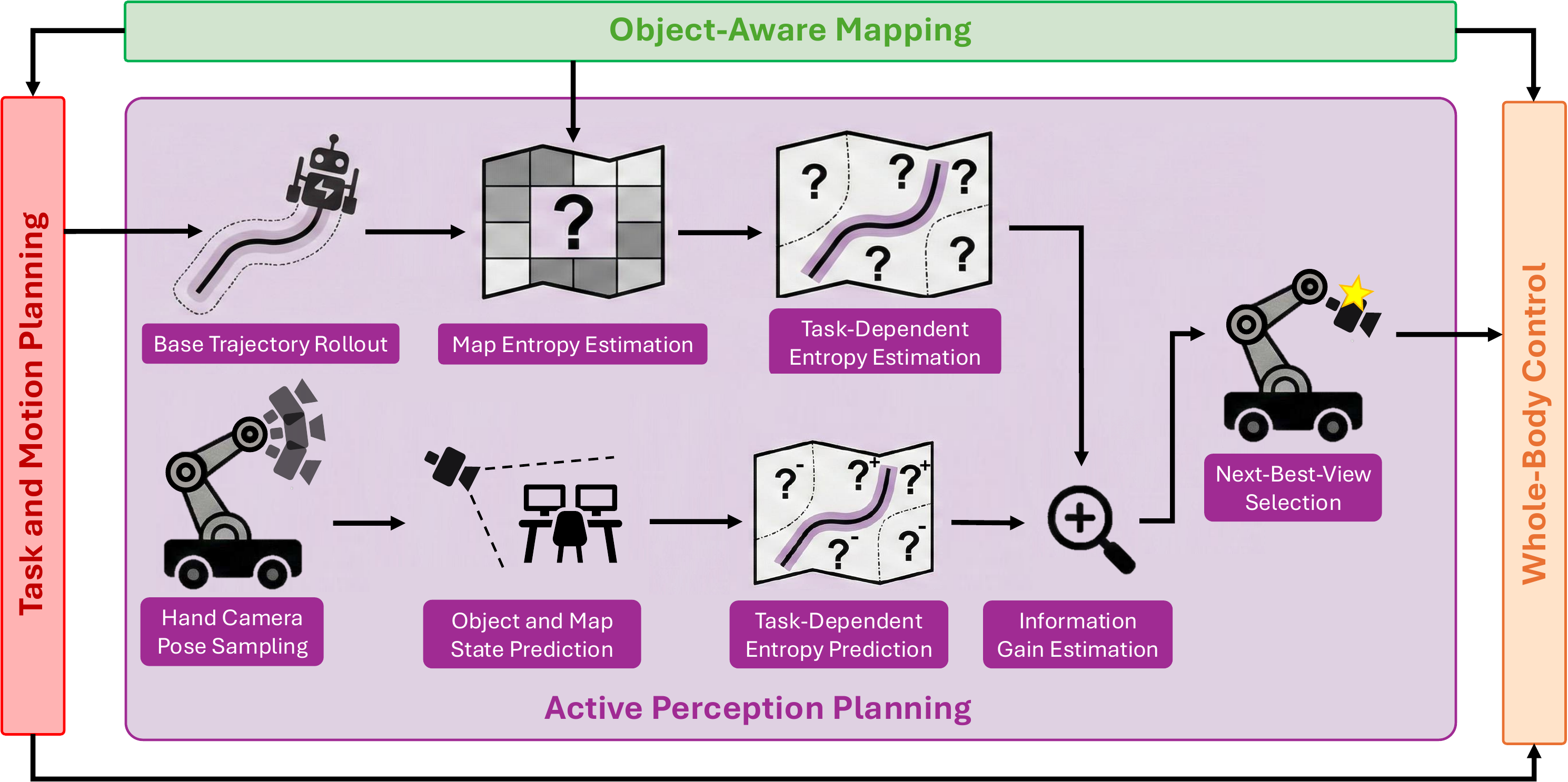}
  \caption{Detailed schematic of the proposed active perception module. \vspace{-4mm}}
\label{fig:system_nbv}
\end{figure*}

The active perception module receives the maintained map, estimated robot state, and desired base path, and selects a hand camera viewpoint predicted to reduce task-dependent map entropy. As illustrated in \autoref{fig:system_nbv}, it ray traces the cells visible from each of $R$ sampled poses, weights them using the predicted base motion along the path, and scores each pose using the approximate EIG defined in \autoref{sec:method-eig}.

Object-level consistency identifies potentially changed regions, while path-dependent weighting prioritizes those relevant to navigation. Together, these components favor observations that can update the map before an upcoming maneuver.

The maintained map incorporates the hand-camera view history $\mathcal{T}_k$. Given $R$ candidate poses $\{\mathbf{T}_{k+1,r}\}_{r=1}^R$, the selected NBV is
\begin{equation}
    \mathbf{T}^d_{k+1}
    =
    \argmax_r \mathrm{EIG}(\mathbf{T}_{k+1,r}),
\end{equation}
where $\mathrm{EIG}(\cdot)$ is the approximate expected information-gain score. The selected pose becomes the hand-camera reference for control. The map uncertainty, path weighting, and viewpoint score are developed in \autoref{sec:method-map}--\autoref{sec:method-eig}.

\vspace{-2mm}

\subsection{Whole-Body Control}\label{sec:system-control}
The whole-body controller receives the desired base path, current robot and map state, and selected hand-camera reference. It coordinates base tracking and camera motion using the ordered task list
\begin{equation}
    [\task_{\mathit{base}}, \task_{\mathit{cam}}],
\end{equation}
where $\task_{\mathit{base}}$ tracks the base path and $\task_{\mathit{cam}}$ targets the NBV pose. HTMPC gives higher priority to base tracking, uses remaining redundancy for camera motion, and enforces map-derived collision constraints. Both references are updated as the map changes. The controller is formulated in \autoref{sec:htmpc}.

\section{Next-Best-View Formulation}\label{sec:method}
This section formalizes the proposed object- and task-aware NBV module. We first construct the maintained map uncertainty and desired-path prior, then derive the task-dependent map entropy, predict its update for each candidate view, and select the NBV using approximate expected information gain. \vspace{-3mm}

\subsection{Map and Uncertainty Construction}
\label{sec:method-map}
The mapping module maintains an object library $\mathcal{O}=\{\mathbf{O}_i\}$ where each object $\mathbf{O}_i$ carries a local EDF submap $\mathcal{M}_i$ and a state model $p(l_i,v_i)$. We merge these submaps into a single 3D EDF used in whole-body control for collision avoidance by taking the minimum EDF value over all object submaps,
\begin{equation}
\label{eq:method-3d-edf}
    \mathcal{M}_{edf}(\cdot)
    =
    \min_{\mathbf{O}_i \in \mathcal{O}} \mathcal{M}_i(\cdot).
\end{equation}
The resulting $\mathcal{M}_{edf}$ is the 3D collision map later interpolated in \autoref{sec:htmpc} for whole-body collision avoidance~\cite{duPerceptiveHierarchicalTaskMPC2026}.

For navigation and NBV, we evaluate this map on a 2D grid $\mathcal{G}=\{(m,n)\}$ defined at the base height $z_b$, with cell centers $\mathbf{c}_{mn}=[x_{mn},y_{mn}]^T$. The sliced EDF value at each cell is
\begin{equation}
\label{eq:method-slice-edf}
    u_s(\mathbf{c}_{mn})
    \defeq
    \mathcal{M}_{edf}(x_{mn},y_{mn},z_b),
    \qquad (m,n)\in\mathcal{G}.
\end{equation}
Bilinear interpolation over this discrete slice then yields the continuous approximation $u_s(\mathbf{p})$ used in the NBV Fisher information derivation. For each slice cell, we define the index of the closest mapped object by
\begin{equation}
\label{eq:method-nearest-obj}
    i_{mn}^*
    \defeq
    \argmin_{\mathbf{O}_i \in \mathcal{O}}
    \mathcal{M}_i(x_{mn},y_{mn},z_b).
\end{equation}
A cell is treated as object-associated when its closest mapped object lies within a distance band $\delta_o$, i.e., when $u_s(\mathbf{c}_{mn}) \le \delta_o$. For these cells, the map uncertainty is inherited from the corresponding object-consistency variance $\mathrm{Var}[v_{i_{mn}^*}]$ in \eqref{eq:bg-consistency-var}. If an object is not re-observed for an extended period, the mapping module gradually penalizes it with synthetic outlier measurements, which lowers the consistency estimate and enlarges this variance without forcing immediate object removal.

The shared object belief couples the uncertainty assigned to its associated cells. For example, observations of a shifted cabinet's visible edge can change the maintained consistency belief for that cabinet, affecting the uncertainty of its associated region even where the surface remains occluded. This gives subsequent viewpoint selection a reason to inspect the rest of the object. Path weighting, introduced next, determines how strongly that region matters to the navigation task.

Cells with $u_s(\mathbf{c}_{mn}) > \delta_o$ are modeled independently by a Beta occupancy model,
\begin{equation}
\label{eq:method-occ-prior}
    o_{mn,k} \sim \mathrm{Beta}(\alpha^{\mathrm{occ}}_{mn,k},\beta^{\mathrm{occ}}_{mn,k}),
\end{equation}
whose parameters are updated from a binary occupancy observation $z^{\mathrm{occ}}_{mn,k}\in\{0,1\}$ whenever the cell is visible:
\begin{equation}
\label{eq:method-occ-update}
\begin{aligned}
    \alpha^{\mathrm{occ}}_{mn,k+1}
    &= \alpha^{\mathrm{occ}}_{mn,k}
    + \eta_{\mathrm{occ}} z^{\mathrm{occ}}_{mn,k}, \\
    \beta^{\mathrm{occ}}_{mn,k+1}
    &= \beta^{\mathrm{occ}}_{mn,k}
    + \eta_{\mathrm{occ}} \left(1 - z^{\mathrm{occ}}_{mn,k}\right),
\end{aligned}
\end{equation}
where $\eta_{\mathrm{occ}}>0$ is an observation weight and $\eta_{\mathrm{occ}}=1$ recovers the standard Beta-Bernoulli update. If a non-object cell is not re-observed, its parameters are relaxed toward an uninformative prior $(\alpha^{\mathrm{occ}}_0,\beta^{\mathrm{occ}}_0)$ to reflect growing uncertainty:
\begin{equation}
\label{eq:method-occ-relax}
\begin{aligned}
    \alpha^{\mathrm{occ}}_{mn,k+1}
    &=
    (1-\rho)\alpha^{\mathrm{occ}}_{mn,k}
    + \rho \alpha^{\mathrm{occ}}_0, \\
    \beta^{\mathrm{occ}}_{mn,k+1}
    &=
    (1-\rho)\beta^{\mathrm{occ}}_{mn,k}
    + \rho \beta^{\mathrm{occ}}_0,
\end{aligned}
\end{equation}
where $\rho\in(0,1)$ is a forgetting factor. Since $o_{mn,k}$ is also Beta distributed, its variance follows the same form as \eqref{eq:bg-consistency-var}, with $(\alpha_{i,k},\beta_{i,k})$ replaced by $(\alpha^{\mathrm{occ}}_{mn,k},\beta^{\mathrm{occ}}_{mn,k})$.

Finally, we define the effective cell-wise EDF variance used in the NBV objective as
\begin{equation}
\label{eq:method-slice-var}
    \sigma_{s,mn}^2
    \defeq
    \begin{cases}
        \lambda_{\mathrm{edf}}\, \mathrm{Var}[v_{i_{mn}^*}],
        & u_s(\mathbf{c}_{mn}) \le \delta_o, \\
        \lambda_{\mathrm{edf}}\, \mathrm{Var}[o_{mn,k}],
        & \text{otherwise},
    \end{cases}
\end{equation}
where $\lambda_{\mathrm{edf}}>0$ is a scaling factor. Note that $\mathrm{Var}[v_i]$ in \eqref{eq:bg-consistency-var} and $\mathrm{Var}[o_{mn,k}]$ describe object consistency and map occupancy uncertainties, not EDF uncertainty, $\mathrm{Var}\!\left[\mathrm{EDF}(\mathbf{c}_{mn})\right]$, directly. Prior work in uncertainty-aware object mapping derives a nonlinear forward propagation from latent occupancy uncertainty to signed-distance variance~\cite{Liao2023Uncertaintyaware3O}. However, the inverse mapping needed here is not available in closed form. We therefore use the latent consistency and occupancy variances as monotone surrogates and interpret \eqref{eq:method-slice-var} as an effective linear EDF-uncertainty model. The merged 3D map provides the distance field for whole-body collision avoidance, while the 2D slice provides the distance field for base path planning together with the uncertainty field used for NBV selection. \vspace{-2mm}

\subsection{Desired Path and Robot Position Prior}
\label{sec:method-prior}
The navigation task is represented by a dense reference trajectory with length $I$:
\begin{equation}
\mathbf{X}^d = \{\mathbf{p}^d_i\}_{i=0,\dots,I},
\end{equation}
where $\mathbf{p}^d_i \in \Real^2$ denotes the desired 2D base position at rollout step $t_i=i\Delta t$ along the reference path. During execution, the robot may deviate from this path because of tracking error, collision avoidance, and replanning. We therefore model the robot position as a stochastic rollout around the reference path. This rollout provides the spatial task-relevance weights for NBV, while HTMPC uses a separate, short prediction horizon for control.

For the holonomic base, let $\mathbf{p}_i = [p_{x,i} ~ p_{y,i}]^T$ denote the 2D rollout position and $\bar{\mathbf{u}}_i = [v_{x,i} ~ v_{y,i} ]^T$ denote the nominal velocity derived from the desired path produced by the global planner. We model the executed velocity as
\begin{equation}
\mathbf{u}_i
=
\bar{\mathbf{u}}_i + \mathbf{w}_i,
\qquad
\mathbf{w}_i \sim \mathcal{N}(\mathbf{0}, \sigma_d^2 \mathbf{I}),
\end{equation}
where $\mathbf{w}_i$ is zero-mean process noise with variance $\sigma_d^2 \mathbf{I}$ that captures tracking deviations and local replanning effects. This yields the base position rollout
\begin{equation}
\mathbf{p}_{i+1} = \mathbf{p}_i + \Delta t \cdot \mathbf{u}_i.
\end{equation}
Under this random-walk approximation, the position covariance evolves as
\begin{equation}
\Sigma_{d,i} = \Sigma_{d,0} + i \cdot (\Delta t^2 \sigma_d^2 \mathbf{I}).
\end{equation}
Accordingly, we discretize a Gaussian position model centered at $\mathbf{p}_i^d$ with covariance $\Sigma_{d,i}$ over the grid and normalize the resulting cell weights:
\begin{equation}
\begin{aligned}
\widetilde{w}_{mn\mid i}
&\defeq \exp \Big(
    -\frac{1}{2}
    (\mathbf{c}_{mn} - \mathbf{p}_i^d)^T
    \Sigma_{d,i}^{-1}
    (\mathbf{c}_{mn} - \mathbf{p}_i^d)
		\Big), \\
p(\mathbf{c}_{mn}\mid t_i)
&\defeq
\frac{\widetilde{w}_{mn\mid i}}
{\sum_{(a,b)\in\mathcal{G}}\widetilde{w}_{ab\mid i}}.
\end{aligned}
\end{equation}
We use a uniform prior $p(t_i)=1/(I+1)$ over the rollout steps $i=0,\dots,I$. This normalized path-conditioned distribution weights map regions by how likely the robot is to occupy them along the planned path, using the same grid $\mathcal{G}$ for every candidate view. \vspace{-2mm}

\subsection{Task-Dependent Map Entropy}
\label{sec:method-entropy}
At time step $t_k$, the accumulated hand camera view history is denoted by
\begin{equation}
\mathcal{T}_k = \{\mathbf{T}_1,\dots,\mathbf{T}_k\}.
\end{equation}
Conditioned on $\mathcal{T}_k$, the maintained 2D EDF slice and its uncertainty define the current map belief used by NBV. For a planar position $\mathbf{p}$ lying in cell $(m,n)$, using the sliced EDF mean in \eqref{eq:method-slice-edf} and the effective EDF variance in \eqref{eq:method-slice-var}, the local distance-to-nearest-obstacle observation is modeled as
\begin{equation}
\begin{aligned}
\mathrm{EDF}(\mathbf{p} \mid \mathcal{T}_k)
    &= u_s(\mathbf{p}) + \eta_{s,mn}, \\
\eta_{s,mn} &\sim \mathcal{N}(0, \sigma_{s,mn}^2),
\end{aligned}
\end{equation}
where $u_s(\mathbf{p})$ is the interpolated EDF mean and $\sigma_{s,mn}^2$ is the cell uncertainty in \eqref{eq:method-slice-var}. The term $\eta_{s,mn}$ denotes the corresponding local Gaussian EDF noise, and the map gradient $\nabla u_s(\mathbf{p})$ is estimated by finite differences over the grid.

In our formulation, the unknown parameter is a potential robot base position $\mathbf{p}$ and the observation is the EDF value at that position. Since we approximate the EDF observation as Gaussian, the obstacle-distance likelihood for a robot base position $\mathbf{p}$ inside cell $(m,n)$ is
\begin{equation}
\begin{aligned}
p(\mathrm{EDF} \mid \mathcal{T}_k, \mathbf{p})
    &= p(\mathrm{EDF} \mid u_s(\mathbf{p}), \sigma_{s,mn}^2) \\
    &\propto \exp \Big(
        -\frac{(\mathrm{EDF} - u_s(\mathbf{p}))^2}
        {2 \sigma_{s,mn}^2}
    \Big).
\end{aligned}
\end{equation}
Ignoring constant terms, the corresponding log-likelihood is
\begin{equation}
\begin{aligned}
\log p(\mathrm{EDF} \mid \mathcal{T}_k, \mathbf{p})
    \sim
    -\frac{1}{2}\sigma_{s,mn}^{-2}
    (\mathrm{EDF} - u_s(\mathbf{p}))^2.
\end{aligned}
\end{equation}
The Fisher score then becomes
\begin{equation}
\nabla \log p(\mathrm{EDF} \mid \mathcal{T}_k, \mathbf{p})
= \sigma_{s,mn}^{-2}
\big(\mathrm{EDF} - u_s(\mathbf{p})\big)\nabla u_s(\mathbf{p}).
\end{equation}
Evaluating the FIM at the center of cell $(m,n)$ yields
\begin{align}
\mathcal{F}(\mathbf{c}_{mn} \mid \mathcal{T}_k)
&= \mathbb{E}_{\mathrm{EDF}} \Big[
    \big(\nabla\log p(\mathrm{EDF} \mid \mathcal{T}_k, \mathbf{c}_{mn})\big)
    \notag\\
&\qquad\qquad \cdot
    \nabla\log p(\mathrm{EDF} \mid \mathcal{T}_k, \mathbf{c}_{mn})^T
\Big] \\
&= \sigma_{s,mn}^{-2}
\nabla u_s(\mathbf{c}_{mn}) \nabla u_s(\mathbf{c}_{mn})^T .
\end{align}
Cells with zero EDF gradient are omitted from entropy evaluation. For the remaining cells, $\mathcal{F}(\mathbf{c}_{mn} \mid \mathcal{T}_k)$ is rank-1 and therefore non-invertible. Intuitively, the robot only gains information about the closest obstacle along the local map-gradient direction $\nabla u_s(\mathbf{c}_{mn})$. Its non-zero eigenvalue is
\begin{equation}
\lambda_{mn} =
\sigma_{s,mn}^{-2}
\lVert \nabla u_s(\mathbf{c}_{mn}) \rVert^2.
\end{equation}
Under a local linearization, $\lambda_{mn}^{-1}$ gives the variance scale along the EDF-gradient direction. The differential entropy of the corresponding 1D Gaussian approximation is
\begin{equation}
\begin{aligned}
h_e^{1D}(\mathbf{c}_{mn}; \mathcal{T}_k)
    = \frac{1}{2}\log \Big[
        2 \pi e \cdot
        \frac{\sigma_{s,mn}^{2}}
        {\lVert \nabla u_s(\mathbf{c}_{mn}) \rVert^2}
    \Big].
\end{aligned}
\end{equation}
From this expression, the 1D entropy along the map-gradient direction increases with larger cell uncertainty $\sigma_{s,mn}^2$ and smaller gradient magnitude $\lVert \nabla u_s(\mathbf{c}_{mn}) \rVert$. Intuitively, a cell is more uncertain either because its state is less certain or because the local EDF varies weakly with position, making a distance observation less informative. In free-space regions with $u_s(\mathbf{c}_{mn}) > \delta_o$, using the same object-association threshold $\delta_o$ introduced in \autoref{sec:method-map}, we further clip the EDF gradient magnitudes from above to avoid overconfidence in previously observed empty space. This increases the entropy assigned to such regions and encourages the robot to re-observe them for newly added obstacles.
At each grid cell, we compute this one-dimensional Fisher-derived entropy from the effective EDF uncertainty and local map gradient. We define task-dependent map entropy as its expectation over the normalized path-conditioned distribution of grid cells and rollout times:
\begin{equation}
\begin{aligned}
&\mathbb{E}_{\mathbf{p}, t}[h_e^{1D}(\mathcal{T}_k)]\\
&= \sum_{i=0}^I \sum_{(m,n) \in \mathcal{G}}
    h_e^{1D}(\mathbf{c}_{mn}; \mathcal{T}_k) 
  p(\mathbf{c}_{mn} \mid t_i)\, p(t_i).
\end{aligned}
\end{equation}
This task-dependent map entropy serves as the prior term for scoring each candidate viewpoint by approximate EIG. \vspace{-3mm}

\subsection{Candidate Views and Visible-Cell Set}
\label{sec:method-view}
At time step $t_k$, let $\mathcal{V}_k \subset \SE{3}$ denote the candidate hand-camera target pose space. The formulation is continuous in principle but we sample a finite candidate set from $\mathcal{V}_k$:
\begin{equation}
\{\mathbf{T}_{k+1,r}\}_{r=1,\dots,R}
\subset \mathcal{V}_k,
\end{equation}
where each $\mathbf{T}_{k+1,r}$ is a candidate hand-camera target for the next observation. Our path-guided implementation is described in \autoref{sec:experiments}. In the controller, camera orientation is enforced more strongly than position, allowing local position adjustment near obstacles while preserving the intended viewing direction.

Each candidate viewpoint captures a subset of grid cells within its field of view. We denote this visible-cell set by
\begin{equation}
\mathcal{C}_{k+1,r} \defeq \mathcal{C}(\mathbf{T}_{k+1,r}),
\end{equation}
which is obtained by ray tracing on the current maintained map to collect the grid cells expected to be visible from viewpoint $\mathbf{T}_{k+1,r}$. Thus, although the prior entropy in \autoref{sec:method-entropy} is defined over the full grid, only cells in $\mathcal{C}_{k+1,r}$ undergo the predicted map update.

\subsection{Predicted Map Update Under a Candidate View}
\label{sec:method-posterior}
Given a candidate viewpoint $\mathbf{T}_{k+1,r}$, we only predict the posterior for the visible cells in $\mathcal{C}_{k+1,r}$. Since multiple visible cells may belong to the same object, we first collect the associated object set induced by the visible object cells,
\begin{equation}
\mathcal{O}_{k+1,r}
\defeq
\left\{
\mathbf{O}_{i_{mn}^*}
\;\middle|\;
\mathbf{c}_{mn}\in\mathcal{C}_{k+1,r},\;
u_s(\mathbf{c}_{mn}) \le \delta_o
\right\}.
\end{equation}
For each object $\mathbf{O}_i \in \mathcal{O}_{k+1,r}$, we synthesize two geometric measurements: an inlier $\Delta_i^{\mathrm{in}}$ to simulate that the object is stationary, and an outlier $\Delta_i^{\mathrm{out}}$ to simulate that the object has changed, with $|\Delta_i^{\mathrm{in}}| \ll |\Delta_i^{\mathrm{out}}|$. Applying the Bayesian update in \eqref{eq:bg-bayes} with these two synthetic measurements yields two candidate posterior consistency variances, denoted by $\widehat{\mathrm{Var}}^{\mathrm{in}}[v_i]$ and $\widehat{\mathrm{Var}}^{\mathrm{out}}[v_i]$. We retain the posterior with the smaller variance and denote it by
\begin{equation}
\widehat{V}^{\mathrm{obj}}_{i}
\defeq
\min\!\left(
\widehat{\mathrm{Var}}^{\mathrm{in}}[v_i],
\widehat{\mathrm{Var}}^{\mathrm{out}}[v_i]
\right).
\end{equation}
This is an optimistic posterior approximation: rather than integrating over all possible future measurements, we evaluate two representative outcomes and use the lower posterior variance as an estimate of the uncertainty reduction achievable by observing the object. All visible cells associated with object $\mathbf{O}_i$ inherit this predicted variance.

Visible non-object cells with $u_s(\mathbf{c}_{mn}) > \delta_o$ are then updated independently. For each such cell, we synthesize two binary occupancy measurements, occupied and free, apply the weighted Beta-Bernoulli update in \eqref{eq:method-occ-update}, and denote the corresponding posterior variances by $\widehat{\mathrm{Var}}^{1}[o_{mn,k}]$ and $\widehat{\mathrm{Var}}^{0}[o_{mn,k}]$. We use the same approximation and retain the smaller one:

\begin{equation}
\widehat{V}^{\mathrm{occ}}_{mn}
\defeq
\min\!\left(
\widehat{\mathrm{Var}}^{1}[o_{mn,k}],
\widehat{\mathrm{Var}}^{0}[o_{mn,k}]
\right).
\end{equation}
For compactness, denote the predicted effective EDF variance under candidate $r$ as $\widehat{\sigma}_{s,mn,r}^2$:
\begin{equation}
\label{eq:method-post-var}
\widehat{\sigma}_{s,mn,r}^2
\defeq
\begin{cases}
\lambda_{\mathrm{edf}} \widehat{V}^{\mathrm{obj}}_{i_{mn}^*}, &
\mathbf{c}_{mn}\in\mathcal{C}_{k+1,r},\;
u_s(\mathbf{c}_{mn}) \le \delta_o, \\[0.8ex]
\lambda_{\mathrm{edf}} \widehat{V}^{\mathrm{occ}}_{mn}, &
\mathbf{c}_{mn}\in\mathcal{C}_{k+1,r},\;
u_s(\mathbf{c}_{mn}) > \delta_o, \\[0.8ex]
\sigma_{s,mn}^2, & \mathbf{c}_{mn}\notin\mathcal{C}_{k+1,r}.
\end{cases}
\end{equation}
To predict candidate-dependent geometry, we mark a thin layer of currently free cells along the candidate FoV frontier as hypothetically occupied and recompute the EDF $\widehat u_{s,r}$ and its gradient within $\mathcal{C}_{k+1,r}$. This geometric hypothesis is applied independently of which synthetic observation yields the smaller variance. Outside the FoV, we set $\widehat u_{s,r}=u_s$, neglecting induced changes there for efficiency. The predicted posterior entropy used in the EIG calculation is
\begin{equation}
h_e^{1D}(\mathbf{c}_{mn}; \mathcal{T}_k, \mathbf{T}_{k+1,r})
=
\frac{1}{2}
\log \Bigg[
2 \pi e \cdot
\frac{\widehat{\sigma}_{s,mn,r}^2}
{\lVert \nabla \widehat u_{s,r}(\mathbf{c}_{mn}) \rVert^2}
\Bigg].
\end{equation}

\subsection{Expected Information Gain and NBV Selection}
\label{sec:method-eig}
\autoref{sec:method-entropy} derives the current task-dependent map entropy, and \autoref{sec:method-posterior} derives its predicted value under each candidate viewpoint. After incorporating candidate viewpoint $\mathbf{T}_{k+1,r}$, the task-dependent map entropy is
\begin{equation}
\begin{aligned}
&\mathbb{E}_{\mathbf{p}, t}[h_e^{1D}(\mathcal{T}_k, \mathbf{T}_{k+1,r})]\\
&= \sum_{i=0}^I \sum_{(m,n) \in \mathcal{G}}
    h_e^{1D}(\mathbf{c}_{mn}; \mathcal{T}_k, \mathbf{T}_{k+1,r}) p(\mathbf{c}_{mn} \mid t_i)\, p(t_i).
\end{aligned}
\end{equation}
The approximate expected information gain (EIG) of candidate viewpoint $\mathbf{T}_{k+1,r}$ is then defined as the reduction from prior to predicted posterior task-dependent map entropy:
\begin{equation}
\begin{aligned}
\mathrm{EIG}(\mathbf{T}_{k+1,r})
    &= \mathbb{E}_{\mathbf{p}, t}[h_e^{1D}(\mathcal{T}_k)] \\
    &\quad - \mathbb{E}_{\mathbf{p}, t}[h_e^{1D}(\mathcal{T}_k, \mathbf{T}_{k+1,r})].
\end{aligned}
\end{equation}
The expectation uses the normalized grid-cell and rollout-time distribution, with the optimistic posterior prediction defined in \autoref{sec:method-posterior}. A larger approximate EIG therefore predicts a greater reduction in task-dependent map entropy. Under the FoV-local approximation, non-visible cells retain their prior map state, so their contributions cancel and only visible cells need to be considered:
\begin{equation}
\begin{aligned}
\mathrm{EIG}(\mathbf{T}_{k+1,r})
&= \sum_{i=0}^I \sum_{(m,n) \in \mathcal{C}_{k+1,r}}
    \Big(
        h_e^{1D}(\mathbf{c}_{mn}; \mathcal{T}_k) \\
& -
        h_e^{1D}(\mathbf{c}_{mn}; \mathcal{T}_k, \mathbf{T}_{k+1,r})
    \Big)
    p(\mathbf{c}_{mn} \mid t_i)\, p(t_i).
\end{aligned}
\end{equation}
Among the candidate set defined in \autoref{sec:method-view}, the next-best viewpoint is finally selected as
\begin{equation}
\mathbf{T}^d_{k+1} = \arg\max_r \mathrm{EIG}(\mathbf{T}_{k+1,r}).
\end{equation}

\section{Hierarchical-Task MPC}\label{sec:htmpc}
The mobile manipulator assigns higher priority to tracking the desired base trajectory than to realizing the NBV hand camera poses. We employ HTMPC to enforce this task ordering and generate smooth whole-body motion for navigation while supporting active perception. The NBV module provides the selected hand-camera pose $\mathbf{T}^d_{k+1}$, which is held constant as the camera reference for a fixed update interval.

We consider a mobile manipulator with motion model
\begin{equation}\label{eq:HTMPC-motion-model}
\dot \x = \A(\q)\x + \B\u,
\end{equation}
where 
\begin{equation}
   \A = \begin{bmatrix}
   \mathbf{0}_{n\times n}& \mathbf{G}(\q)\\
   \mathbf{0}_{m\times n}&\mathbf{0}_{m\times m}
   \end{bmatrix},\;
    \B = \begin{bmatrix}
   \mathbf{0}_{n\times m}\\
   \mathbf{I}_{m\times m}
   \end{bmatrix}.
\end{equation}
The generalized coordinate $\q$ and velocity $\v$ are related by $\dot \q = \mathbf{G}(\q) \v$, where $\mathbf{G}(\q)$ is a block diagonal matrix with two components: an identity matrix for the arm and a position-dependent matrix for the mobile base~\cite{siciliano-robotics-mobile-modeling}. We choose accelerations as the control inputs, $\u = \dot \v$. If the base is holonomic, the matrix $\mathbf{G}(\q)$ degenerates to an identity matrix $\mathbf{G} = \mathbf{I}_{n \times n}$. As a result, the holonomic mobile manipulator can be conveniently modeled as a double integrator.

Given the ordered task list $[\task_{\mathit{base}},\task_{\mathit{cam}}]$, HTMPC assigns higher priority to $\mathcal{J}_{\scriptstyle base}$ than to $\mathcal{J}_{\scriptstyle cam}$. Both tasks are formulated as pose-tracking tasks whose cost is the accumulated tracking error $\mathbf{e}$ over the prediction horizon $[t_k, t_k +T_{\mathrm{mpc}}]$:
\begin{equation} \label{eq:HT-MPC-cont-cost-scalar}
    \mathcal{J} = \frac{1}{2} \int_{t_k}^{t_k+T_{\mathrm{mpc}}}\pnorm{\mathbf{e}}{\Q_e}^2 \,dt,
\end{equation}
where $\Q_{e} \in \Real^{6 \times 6}$ is a positive-definite diagonal weighting matrix and can be used to trade off the position and orientation error. In practice, the $\Q_{e}$ matrix can be designed to penalize the hand camera orientation error more since small variations in view angle often produce more changes in visibility and information gain than small translations.   
The error vector~$\mathbf{e}$ is given by
\begin{equation}\label{eq:pose_tracking_task_distance}
    \mathbf{e} = \begin{bmatrix}
                    \mathbf{e}_{pos} \\
                    \mathbf{e}_{orn}
                 \end{bmatrix}
                 = \begin{bmatrix}
                    \pt^d - \pt \\
                    \Log(\rot^{-1} \rot^d)^\vee
                 \end{bmatrix}
\end{equation}
with position error $\mathbf{e}_{pos}$ and orientation error $\mathbf{e}_{orn}$. The orientation error is computed using the matrix logarithm $\Log: \SO{3} \rightarrow \so{3}$ followed by the vee operator $(\cdot)^\vee:\so{3}\rightarrow\Real^3$. The desired position $\pt^d$ and orientation $\rot^d$ come from the planned navigation path $\mathbf{X}^d$ for the base task and from the selected NBV hand-camera pose $\mathbf{T}^d_{k+1}$ for the camera task. 

\begin{figure*}[!t]
    \centering
    \includegraphics[width=0.99\linewidth]{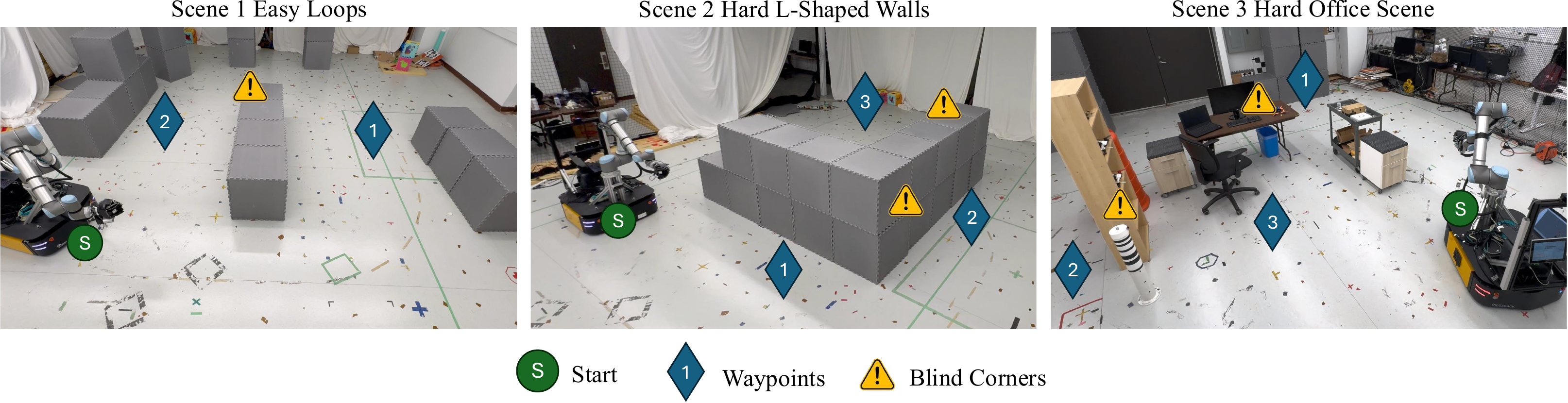}
    \caption{Real-world static-scene benchmark. Left: Scene 1 Easy Loops. Middle: Scene 2 Hard L-Shaped Walls. Right: Scene 3 Hard Office Scene. The scenes progress from open loops to sharper turns and cluttered occlusions, where hidden hazards near the waypoints make active map maintenance important.\vspace{-3mm}}
    \label{fig:real_world_scenes}
\end{figure*}

Environment and self-collision avoidance are explicitly incorporated as the nonlinear constraints in \eqref{eq:bg-htmpc-nonlinear}.
To approximate the robot's envelope, we cover the robot arm and base with a set of $M$ collision spheres and use them as the collision model. Each collision sphere $S_j$ has radius~$r_j$ and center $\pt^{S_j}(\q)$ obtained from forward kinematics.

For environment collision avoidance, we impose a control-barrier-function (CBF)-based constraint for each collision sphere $S_j$. We define the collision-free safe set so that each sphere maintains a desired obstacle clearance $\delta_{\mathit{safe}}$:
\begin{equation}
\label{eq:map_to_cbf}
    \begin{aligned}
    h_j(\x)
    &= \textrm{Interp3D}(\mathcal{M}_{edf})(\pt^{S_j}(\q))
    - r_j - \delta_{\mathit{safe}}, \\
    &\hspace{7em} \forall j = 1, 2, \dots, M,
    \end{aligned}
\end{equation}
where $\mathcal{M}_{edf}$ is the 3D EDF map constructed from the object library in \autoref{sec:method-map}, and $\textrm{Interp3D}$ denotes trilinear interpolation, making $h_j$ continuous and piecewise differentiable. The CBF-based safety constraints are then
\begin{equation}
\label{eq:env_collision_avoidance_cbf}
    \dot{h}_j(\x, \u) > -\gamma h_j(\x), \quad \forall j = 1, 2, \dots, M.
\end{equation}
Equation \eqref{eq:env_collision_avoidance_cbf} is imposed as a clearance-rate constraint along the MPC trajectory. Intuitively, these constraints limit how quickly the robot can move toward obstacles. The constant $\gamma$ acts as a decay factor that shapes this speed limit and the resulting traversal clearance. A larger $\gamma$ allows the robot to behave more aggressively around obstacles, while a smaller $\gamma$ enforces more conservative behavior.
For self-collision avoidance, let $\mathcal{P}_{\mathit{self}}$ denote the selected set of nonadjacent collision-sphere pairs that can collide under the robot's joint limits. We impose a Euclidean-distance constraint for each pair:
\begin{equation}
\label{eq:self_collision_avoidance_edf}
    \left\lVert \pt^{S_i}(\q) - \pt^{S_j}(\q) \right\rVert_2 - r_i - r_j -\delta_{\mathit{safe}} \geqslant 0,
    \quad \forall(i,j)\in\mathcal{P}_{\mathit{self}}.
\end{equation}

Both EDF- and CBF-based formulations are effective in idealized settings. In practice, however, the online constructed map is affected by perception delays, occlusions, and uncertainty, which are difficult to model explicitly in planning and control. We therefore choose the CBF formulation for environment collision avoidance and use the extended class function, through the parameter $\gamma$, to define more conservative behavior around obstacles~\cite{QianMPCCBF, duPerceptiveHierarchicalTaskMPC2026}.

Collecting the cost functions and constraints defined above, the ideal lexicographic HTMPC problem is written as
\begin{equation}\label{eq:mth-htmpc}
\begin{aligned}
      \lexmin_{\x,\u}\;& [\mathcal{J}_{\scriptstyle base},\mathcal{J}_{\scriptstyle cam}] \\
      \text{s.t.}\;& \eqref{eq:HTMPC-motion-model},\eqref{eq:env_collision_avoidance_cbf}, \eqref{eq:self_collision_avoidance_edf},\eqref{eq:bg-htmpc-xu-cst},\eqref{eq:bg-htmpc-init-cst}.
\end{aligned}
\end{equation}
The base and camera objectives are minimized in lexicographic order subject to the motion model \eqref{eq:HTMPC-motion-model}, safety constraints \eqref{eq:env_collision_avoidance_cbf} and \eqref{eq:self_collision_avoidance_edf}, state and input constraints \eqref{eq:bg-htmpc-xu-cst}, and initial conditions \eqref{eq:bg-htmpc-init-cst}. 
The HTMPC controller is solved in a receding-horizon fashion with the latest robot state, maintained map, planned base reference, and hand camera reference. In practice, we approximate the ideal hierarchy with two consecutive regularized STMPCs, preserving the base-tracking cost attained in the first stage. The first STMPC computes a numerically regularized base-tracking solution:
\begin{equation}
\label{eq:htmpc-base-stmpc}
\begin{aligned}
(\x_{\mathit{base}}^*, \u_{\mathit{base}}^*)
&= \argmin_{\x,\u}\;
\mathcal{J}_{\mathit{base}} + \mathcal{J}_{\mathit{aux}} \\
\mathrm{s.t.}\quad
& \eqref{eq:HTMPC-motion-model},\eqref{eq:env_collision_avoidance_cbf}, \eqref{eq:self_collision_avoidance_edf},\eqref{eq:bg-htmpc-xu-cst},\eqref{eq:bg-htmpc-init-cst}.
\end{aligned}
\end{equation}
The second STMPC optimizes the hand-camera task subject to an additional task-priority inequality that bounds the base-tracking cost by the value attained in the first stage:
\begin{equation}
\label{eq:htmpc-camera-stmpc}
\begin{aligned}
(\x^*, \u^*)
&= \argmin_{\x,\u}\;
\mathcal{J}_{\mathit{cam}} + \mathcal{J}_{\mathit{aux}} \\
\mathrm{s.t.}\quad
& \eqref{eq:HTMPC-motion-model},\eqref{eq:env_collision_avoidance_cbf}, \eqref{eq:self_collision_avoidance_edf},\eqref{eq:bg-htmpc-xu-cst},\eqref{eq:bg-htmpc-init-cst}\\
& \mathcal{J}_{\mathit{base}}(\x,\u)
\le
\mathcal{J}_{\mathit{base}}(\x_{\mathit{base}}^*, \u_{\mathit{base}}^*).
\end{aligned}
\end{equation}

As a result of this inequality constraint, the camera task is optimized only within solutions that satisfy the attained base-tracking cost bound, encouraging the robot to use its remaining DoFs for active perception. 

An auxiliary cost $\mathcal{J}_{\mathit{aux}}$ is added to both STMPC iterations to improve tracking performance and smoothness by penalizing velocity tracking error and state and control inputs:
\begin{equation}\label{eq:HT-MPC-cont-aux-cost-scalar}
    \mathcal{J}_{\mathit{aux}} = \frac{1}{2} \int_{t_k}^{t_k+T_{\mathrm{mpc}}}\pnorm{\dot{\mathbf{e}}}{\Q_{\dot e}}^2 + \pnorm{\x}{\Q_{x}}^2 + \pnorm{\u}{\Q_{u}}^2 \,dt,
\end{equation}
where $\Q_{\dot e}$, $\Q_{x}$, and $\Q_{u}$ are positive definite weighting matrices of appropriate dimensions.

\begin{figure*}[!t]
    \centering
    \includegraphics[width=0.96\linewidth]{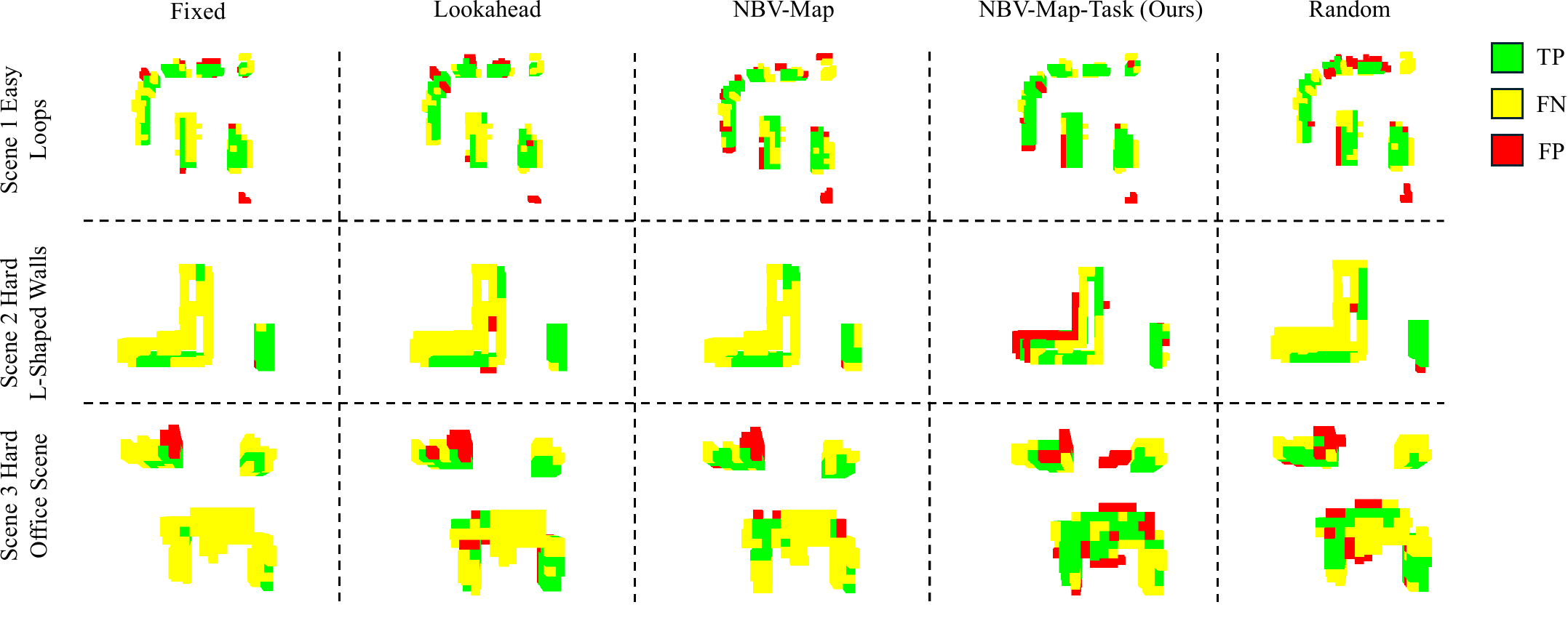}
    \caption{Bird's-eye-view 3D reconstruction results in the real-world static-scene benchmark. Rows correspond to Scene 1--3. Green, yellow, and red indicate true positives, false negatives, and false positives, respectively. The proposed method recovers more occupied voxels in blind and path-critical regions. \vspace{-2mm }}
    \label{fig:map_benchmark}
\end{figure*}

\section{Experiments}\label{sec:experiments}

\subsection{Experimental Setup}

\subsubsection{Platform and System Configuration}
The mobile manipulator platform and overall software architecture are introduced in \autoref{sec:system}, and here we summarize the experiment-specific settings. The perception and mapping stack builds on ORB-SLAM3~\cite{ORBSLAM3_TRO} and POCD~\cite{QianChatrathPOCD}. A modified  ORB-SLAM3 uses the base camera and wheel odometry to estimate the robot's state at 50 Hz, while dense object mapping integrates both base and hand camera observations at $5$--$10$ Hz in parallel threads. We use a uniform semantic label and set the scene-change prior to $s_i = 0$ for all objects, so every mapped object is treated as potentially changing. In the NBV module, $\lambda_{\mathrm{edf}}=10$, $\delta_o=1.8$ m, and the EDF gradient magnitude is clipped to $0.8$ in cells with $u_s(\mathbf{c}_{mn})>\delta_o$. Although the formulation allows sampling from the continuous reachable hand-camera pose space, our implementation uses a path-guided discrete set for runtime: a look-ahead waypoint defines $\psi_k^{\mathrm{ref}}$, and we sample yaw offsets $\Delta\Psi=\{-\pi/3,-\pi/6,0,\pi/6,\pi/3\}$ with reaches $\mathcal{R}=\{1.5,2.0\}$ m from the base center. Candidate reaches that fall too close to obstacles in the sliced EDF are shortened to preserve the safety margin. HTMPC is implemented in ACADOS~\cite{verschueren_acadosmodularopensource_2022} with $T_{\mathrm{mpc}}=3$ s, $\gamma=0.7$, arm collision-sphere radius $r_j=0.25$ m, and obstacle safety margin $\delta_{\mathit{safe}}=0.15$ m. Both controllers run at $10$ Hz, and the final commands are sent at $100$ Hz. Unless otherwise noted, method parameters are fixed across experiments.

\begin{table}[t!]
\scriptsize
\centering
\caption{Quantitative mapping results from the real-world static-scene benchmark. The signed improvement compares ours with the best baseline: ours minus baseline for precision and recall, and baseline minus ours for FPR; positive values favor ours.}
\resizebox{\columnwidth}{!}{
\begin{tabular}{c|cccc}
\Xhline{2\arrayrulewidth}  \Xhline{2\arrayrulewidth}
\textbf{Scene 1 Easy Loops} & \textbf{Precision $\uparrow$} & \textbf{Recall (TPR) $\uparrow$} & \textbf{FPR $\downarrow$} & \textbf{Outcome} \\ \hline
Fixed         & 81.0   &  56.1  & 2.2 & Success  \\
\rowcolor{LightCyan}
Lookahead    &  81.7  &  59.6 & 2.3 & Success \\
NBV-Map    & 81.8 & 61.9   & 2.2   & Success  \\
\rowcolor{LightCyan}
Random &  77.2  &  56.2  &  2.8 & Success  \\
\hline
\textbf{NBV-Map-Task (ours)}    &  \textbf{87.4}  & \textbf{70.9}  & \textbf{1.7} & Success  \\
    \textit{Signed improvement}      & 5.6 & 9.0 & 0.5 & \\\hline
\Xhline{2\arrayrulewidth} \Xhline{2\arrayrulewidth}
\textbf{Scene 2 Hard L-Shaped Walls} & \textbf{Precision $\uparrow$} & \textbf{Recall (TPR) $\uparrow$} & \textbf{FPR $\downarrow$} & \textbf{Outcome} \\ \hline
Fixed         & \textbf{97.6}    &   30.6  & \textbf{0.3}    & Collision$\times$3\\
\rowcolor{LightCyan}
Lookahead       &  89.9  & 31.2   & 1.6 & Collision$\times$3\\
NBV-Map     &  97.4  &  28.4  &   \textbf{0.3}  & Collision$\times$3\\
\rowcolor{LightCyan}
Random  &  94.0  &  30.4  &   0.9  & Collision$\times$3 \\
\hline
\textbf{NBV-Map-Task (ours)}    &  71.4  &  \textbf{73.0}  &  8.4 & Success \\
\textit{Signed improvement}  & -26.2 & 41.8 & -8.1 \\\hline
\Xhline{2\arrayrulewidth} \Xhline{2\arrayrulewidth}
\textbf{Scene 3 Hard Office Scene} & \textbf{Precision $\uparrow$} & \textbf{Recall (TPR) $\uparrow$} & \textbf{FPR $\downarrow$} & \textbf{Outcome} \\ \hline
Fixed         & \textbf{83.3}    &   25.1  & \textbf{1.8}  & Collision$\times$3  \\
\rowcolor{LightCyan}
Lookahead       &  73.1  & 47.5   & 4.4 & Success\\
NBV-Map     &  80.0  &  37.9  &   3.4  & Collision$\times$3\\
\rowcolor{LightCyan}
Random  &  78.7  &  58.3  &   5.4  & Collision$\times$3\\
\hline
\textbf{NBV-Map-Task (ours)}    &  74.4  &  \textbf{70.0} &  8.6 & Success\\
\textit{Signed improvement}  & -8.9 & 11.7 & -6.8 \\\hline
\end{tabular}}
\label{tab:benchmap_quant}
\end{table}

\subsubsection{Baseline Methods}
We compare the proposed object- and task-aware framework against four passive or partially informed baselines that differ only in how the hand camera pose is computed:
\begin{itemize}
    \item \emph{Fixed}: the hand camera is kept at a fixed position and points forward along the base's heading. This is a common default configuration for mobile manipulators and serves as a passive baseline that does not adapt to the scene or task~\cite{QianMPCCBF}.
    \item \emph{Random}: the hand camera is selected uniformly at random from the same discrete candidate view set used by our method. This provides a fair active-perception baseline with the same motion range and view discretization, but without map or task awareness.
    \item \emph{Lookahead}: the hand camera points toward a look-ahead point $8$ s ahead along the desired path~\cite{duPerceptiveHierarchicalTaskMPC2026}. This baseline exploits a simple task-aware heuristic without reasoning about the maintained map state and uncertainty.
    \item \emph{NBV-Map}: the hand camera is selected by the same information-theoretic NBV objective as our method, but without the task-weighting term. This baseline follows the more typical map exploration strategy of selecting views from maintained map uncertainty alone, without prioritizing regions relevant to navigation~\cite{Bircher2016RecedingNBV,Naazare2022OnlineNP}.
\end{itemize}
We denote the full method by \emph{NBV-Map-Task}. Together, the baselines separate passive sensing (Fixed), camera reorientation without map or task reasoning (Random), path-directed sensing (Lookahead), and uncertainty-driven sensing without task weighting (NBV-Map). The comparisons assess how combining map uncertainty and task relevance changes the observations acquired during navigation.

\begin{figure*}[!t]
    \centering
    \includegraphics[width=\linewidth]{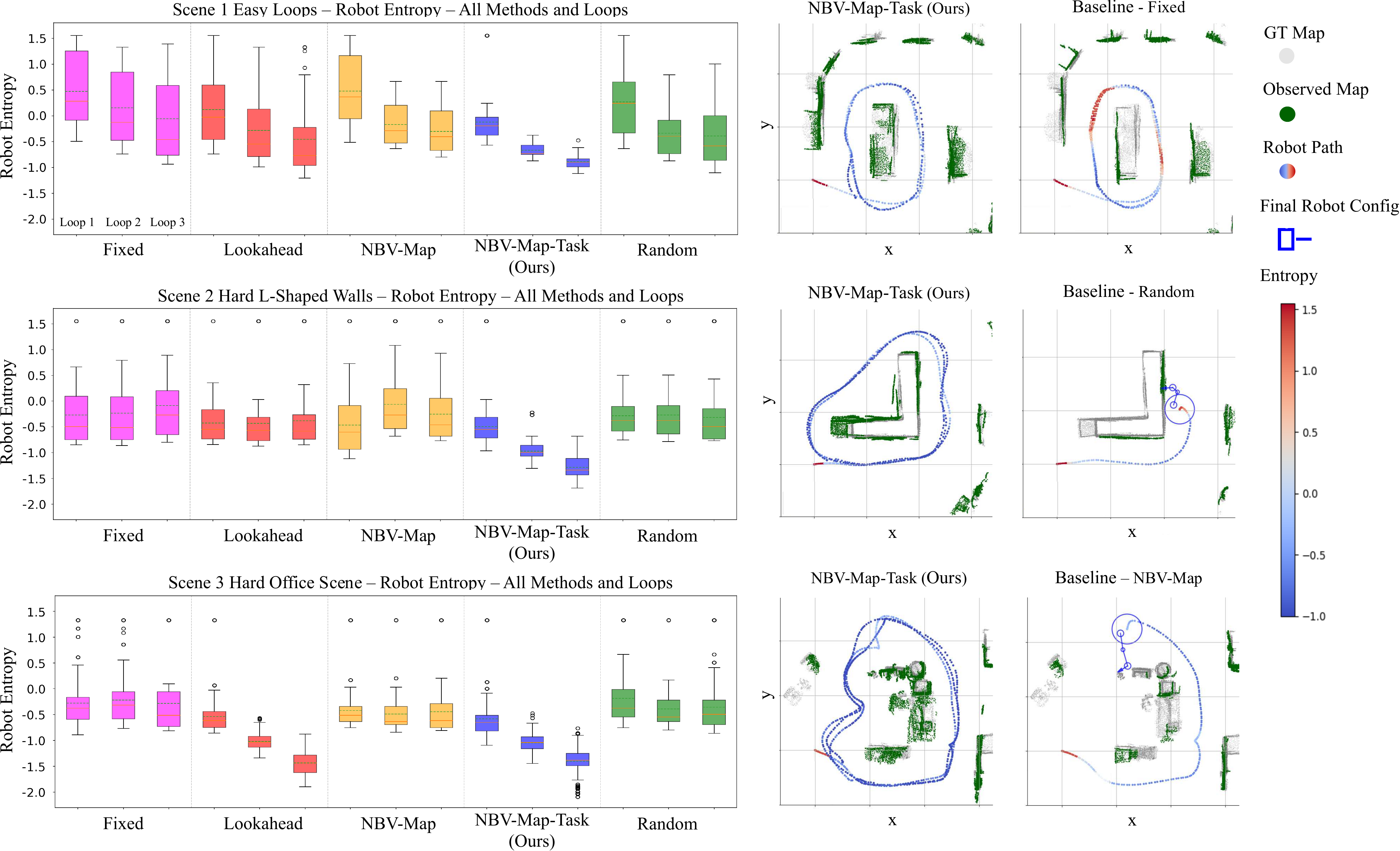}
    \caption{Task-dependent map entropy along executed paths in the real-world static-scene benchmark. Left: boxplots over the three loops in each scene. Right: representative top-down views with trajectories colored by the local entropy. Lower values indicate lower entropy in the regions traversed by the robot. \vspace{-0.2cm}}
    \label{fig:entropy_results}
\end{figure*}

\subsubsection{Experiment Protocol}
We evaluate our framework across three experiment families. The static benchmark isolates viewpoint selection under occlusion. The controlled simulation then tests whether moved objects redirect sensing when the navigation task stays the same. Finally, repeated real-world visits examine whether those map updates support adaptive navigation under scene changes. Together, these experiments progress from acquiring missing geometry to revising previously acquired scene knowledge:
\begin{itemize}
    \item \textbf{Real-world static-scene benchmark}: we conduct a benchmark in the three static, cluttered scenes shown in \autoref{fig:real_world_scenes}: Scene~1 Easy Loops, Scene~2 Hard L-Shaped Walls, and Scene~3 Hard Office Scene. Each method is executed for three consecutive loops in each scene to evaluate mapping quality, task-dependent map entropy, and navigation safety under repeated visits. The first loop starts with an empty map. If a loop succeeds without collision, the maintained map is carried over and updated in the next loop. Collisions trigger a robot and map reset.
    \item \textbf{Controlled scene-change simulation}: we use a controlled Isaac Sim simulation in which objects are shifted between loops to demonstrate how the object-aware uncertainty model qualitatively and quantitatively responds to object-level scene changes.
    \item \textbf{Long-term real-world scene-change experiment}: we perform a long-term real-world experiment in the office scene where a large shelf is relocated between visits to demonstrate incremental map maintenance and adaptive navigation behavior in a changing environment.
\end{itemize}
The multimedia attachment and \href{https://tinyurl.com/nbv-for-mm-nav}{shared video folder} provide videos of the experiments.

\subsection{Real-World Experiments in Static Scenes}

\subsubsection{Scene Overview and Evaluation Metrics}
In this subsection, we evaluate the proposed system in the three real-world scenes shown in \autoref{fig:real_world_scenes}. This benchmark assesses whether the robot can reduce uncertainty and reliably detect obstacles in blind and path-critical regions before executing a turn or passing through a narrow space. The three scenes contain blind corners, narrow passages, and occluded obstacles along the planned path. With passive or task-unaware hand camera motion, hidden obstacles may remain unobserved until the robot has already entered the risky region. In contrast, the proposed method directs the hand camera toward these regions earlier and incorporates the observations before the maneuver.

In all scenes, the robot receives only the blue waypoints shown in \autoref{fig:real_world_scenes} and plans its path online from the maintained map. Yellow exclamation marks identify representative hidden hazards. Scene~1 Easy Loops provides an open setting, while Scene~2 Hard L-Shaped Walls and Scene~3 Hard Office Scene contain sharper turns and stronger occlusions. Starting without a map, the robot may initially plan through unseen obstacles. Successful execution therefore requires observing them before entering the turn.

We report this benchmark in three parts:

\begin{itemize}
    \item \textbf{Map reconstruction quality}: precision, recall, and false positive rate on voxelized maps, after the final loop.
    \item \textbf{Task-dependent map entropy}: evaluated along the executed base trajectory and reported over all loops.
    \item \textbf{Navigation safety}: success rate and distance to the nearest reference-map obstacle, aggregated over all loops.
\end{itemize}

For each real-world scene, we construct a reference map by manually driving the robot to cover as much of the environment as possible. Both the maintained and reference 3D point clouds are voxelized at $0.3$ m resolution. Evaluation is performed over the union of the fixed reference map and each method's reconstructed map. Precision measures how many reconstructed occupied voxels agree with the reference map, recall (TPR) measures how many reference occupied voxels are recovered, and FPR measures how many false occupied voxels appear relative to the reference map. High recall is especially important for navigation because missed occupied voxels can cause the robot to plan through unobserved obstacles. High precision and low FPR indicate fewer spurious occupied voxels, but are not sufficient if large obstacle regions remain unseen. Task-dependent map entropy is evaluated along the executed base trajectory in every loop. Clearance uses the same collision-sphere model as control: we evaluate the distance from each joint sphere surface to the nearest point in the reference point cloud and take the minimum over all joints. Reconstruction is evaluated after the final loop. Entropy, success, and clearance include all loop executions.

\begin{figure*}[!t]
    \centering
    \includegraphics[width=0.99\linewidth]{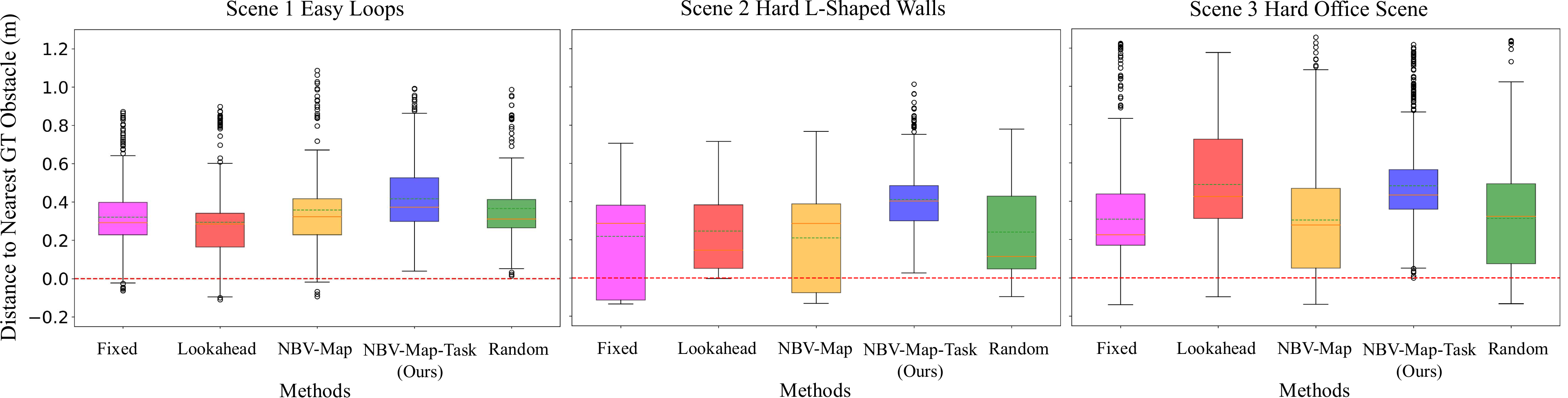}
    \caption{Safety margin in the real-world static-scene benchmark. Boxplots show the point-cloud clearance proxy over all loop executions. The dashed red line marks zero clearance under this proxy, corresponding to a near-contact configuration. Larger positive values indicate greater estimated clearance, especially in the hard turning scenes. \vspace{-0.3cm}}
    \label{fig:distance_to_obstacle}
\end{figure*}

\begin{table}[t!]
\scriptsize
\centering
\caption{Completion percentage across nine sequential loop executions per method: three loops in each of three scenes.}
\begin{tabular}{c|c}
\Xhline{2\arrayrulewidth}  \Xhline{2\arrayrulewidth}
\textbf{Method} & \textbf{Success Rate $\uparrow$}  \\ \hline
Fixed         & 33.3\%    \\
\rowcolor{LightCyan}
Lookahead    &  66.7\% \\
NBV-Map    & 33.3\%  \\
\rowcolor{LightCyan}
Random &  33.3\%  \\
\hline
\textbf{NBV-Map-Task (ours)}    &  \textbf{100.0\%}\\\hline
\end{tabular}
\label{tab:realworld_success_rate}
\end{table}

\subsubsection{Map Reconstruction Quality}
We first compare the voxel maps reconstructed at the end of the final traversal of each scene. This comparison assesses the reconstruction of occupied regions that determine path feasibility over repeated traversals, with map resets after collisions as described above. \autoref{fig:map_benchmark} shows the reconstructions, with voxel metrics in \autoref{tab:benchmap_quant}.

In the easier Scene~1 setup, all methods complete the task, and the proposed method achieves the best precision, recall, and FPR. The recall gap becomes larger in Scene~2 and Scene~3, where critical obstacles lie behind blind turns and within narrow passages. Baselines that do not combine map information and task relevance can achieve high precision and low FPR mainly because they leave substantial parts of the occluded obstacles unobserved, avoiding false occupied voxels in unseen regions at the cost of low recall. Several baseline loops end in collisions near these unobserved regions. This is especially noticeable in Scene~2, where the baseline maps leave large false-negative gaps inside the L-shaped wall, while the proposed method recovers the occupied voxels behind the corner and is the only method that completes the scene. In Scene~3, the proposed map also captures more of the cluttered occupied voxels that support safe turns.

Fixed and Random do not reliably inspect hidden obstacles before reaching a turn. Lookahead improves coverage when the path direction exposes them, but does not explicitly target uncertain map regions. NBV-Map often selects outward frontier regions, whereas the proposed method directs observations toward occupied structures relevant to the next maneuver. This increased coverage comes with lower precision and higher FPR in the hard scenes, where localization drift and sensor noise introduce spurious occupied voxels. The results therefore show a tradeoff between recovering previously hidden obstacles and reconstruction precision.

\subsubsection{Task-Dependent Map Entropy}
We next evaluate whether the selected views reduce uncertainty in the map regions where the robot needs confidence for safe navigation. This metric is aligned with the proposed formulation: map awareness determines where uncertainty remains, and task weighting determines its relevance to the upcoming maneuver. \autoref{fig:entropy_results} reports task-dependent map entropy along the executed base trajectories in every loop. The left panels summarize the distributions over three loops per scene, and the right panels show representative paths colored by local entropy and the final robot configuration when a loop ends in collision.

In Scene~1, all methods complete the task and reduce entropy over repeated loops, while the proposed method achieves lower mean and variance. It also yields lower mean and variance in Scene~2 and Scene~3, where several baselines collide before completing a loop. Failed executions terminate early and reset the map, so they do not accumulate the same uncertainty reduction across traversals. The stronger separation in these scenes is consistent with the reconstruction results. Lookahead helps when the path direction exposes the relevant region, but can leave an occluded turning region unobserved. NBV-Map can instead favor uncertain outward frontiers even when they have little influence on the next maneuver. Task weighting directs the proposed policy toward uncertainty near the upcoming turn, where an additional observation can affect the planned motion.

The right panels of \autoref{fig:entropy_results} show the same relation spatially. In Scene~1, all methods complete the task, but the Fixed baseline still leaves the observed map incomplete around the inner turn, visible as a mismatch between the observed map (dark green) and reference map (gray). The corresponding trajectory passes through higher-entropy cells near the blind corner, showing that the entropy metric reflects local map quality even when the loop succeeds. In Scene~2 and Scene~3, this effect becomes failure-critical. Random leaves the critical turning region unobserved, so the robot plans through the incomplete map toward the next waypoint and collides. NBV-Map shows the complementary failure mode: maintained-map uncertainty is available, but without task weighting the critical turn remains insufficiently observed. In both cases, the final robot configuration lies near the unobserved obstacle region, and the trajectory color shows higher map entropy in that same area. The proposed method observes these regions earlier, so the observed map better matches the reference map and entropy is reduced before the base reaches the turn. Thus, entropy along the path is a representative measure of the local map state that affects navigation safety.

\begin{figure}[b]
\centering
\includegraphics[width=\linewidth]{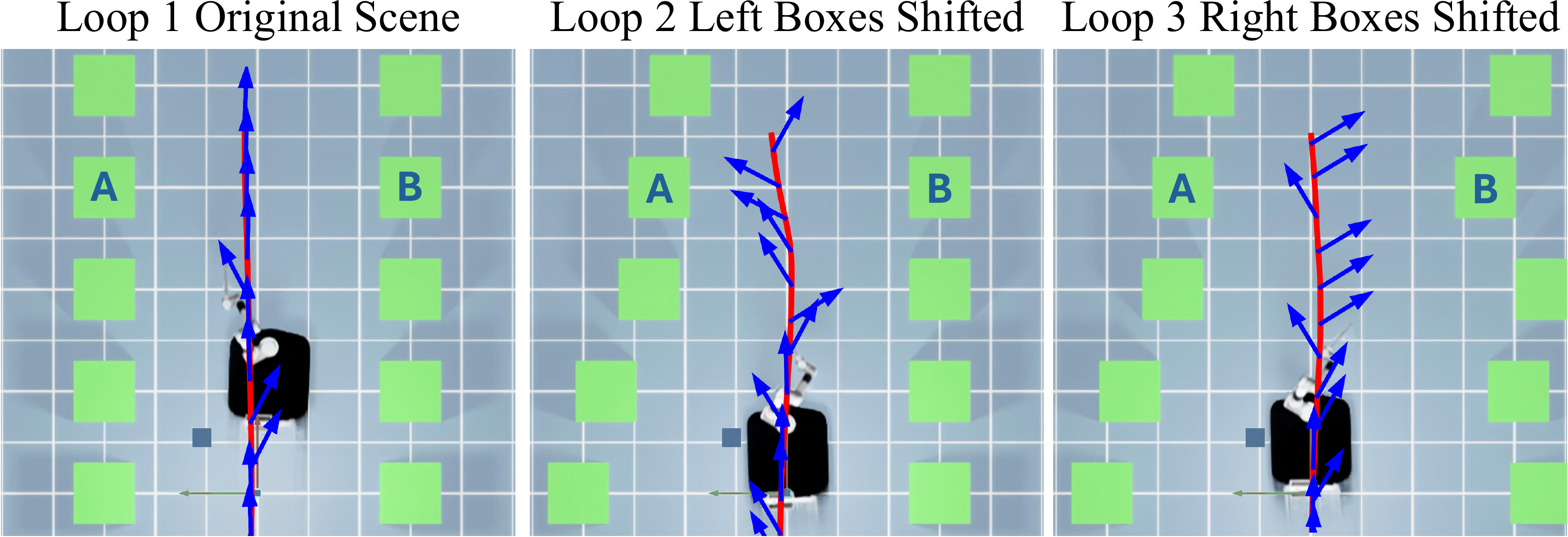}
\caption{Controlled Isaac Sim scene-change experiment. The red curves are the executed base trajectories, and the blue arrows show the sampled hand camera headings. After one row of boxes is shifted, the hand camera redirects toward that side to acquire additional observations.}
\label{fig:semi_static_qual}
\end{figure}

\subsubsection{Navigation Safety}
\autoref{tab:realworld_success_rate} reports the navigation safety benchmark outcomes. The proposed method is the only one that completes all tested loops, whereas each baseline fails in at least one harder scene. \autoref{fig:qual_against_baseline} provides representative examples: the red curve is the planned base path, the blue shaded region is the hand camera FoV, and the yellow marker indicates the hidden hazard. In the blind-corner example, the proposed method turns the hand camera toward the corner before the base reaches the turn, so the map can be updated before the robot commits to the maneuver. The random baseline leaves the same region outside the FoV until the base is already close, so the planner continues to rely on an incomplete local map. The office-scene example shows the same behavior around a hidden pole near the turn. These examples show how map- and task-aware viewpoint selection directs the hand camera toward uncertain regions relevant to the next maneuver to enable timely map updates and support safer navigation.

The accompanying video also shows how the selected views change after these hazards have been inspected. Along the straight segment beyond the L-shaped turn, the camera looks outward toward remaining uncertain regions. In the office, it scans the room after checking the hidden pole. Task-aware sensing therefore adapts throughout the maneuver rather than fixing the camera toward one region.

\autoref{fig:distance_to_obstacle} shows the same effect quantitatively. The dashed red line marks zero clearance based on the controller's collision model. In Scene~1, the methods are more comparable, although several baselines still approach or cross the zero-clearance boundary; the proposed method maintains one of the highest and most consistent clearance distributions. The separation is clearer in Scene~2, where early observation of the blind turn matters most: the proposed method shifts the distribution upward, while the baselines remain closer to near-contact. In Scene~3, Lookahead completes the loops, but its clearance distribution still contains many near-collision events, indicating little safety margin. The proposed method remains more consistent because it selects views using both the maintained map and the upcoming maneuver. These clearance distributions agree with the qualitative examples and uncertainty analysis.

\subsection{Controlled Simulation Under Scene Changes}

\begin{figure}[t!]
    \centering
    \begin{subfigure}[b]{\columnwidth}
        \centering
        \includegraphics[width=\textwidth]{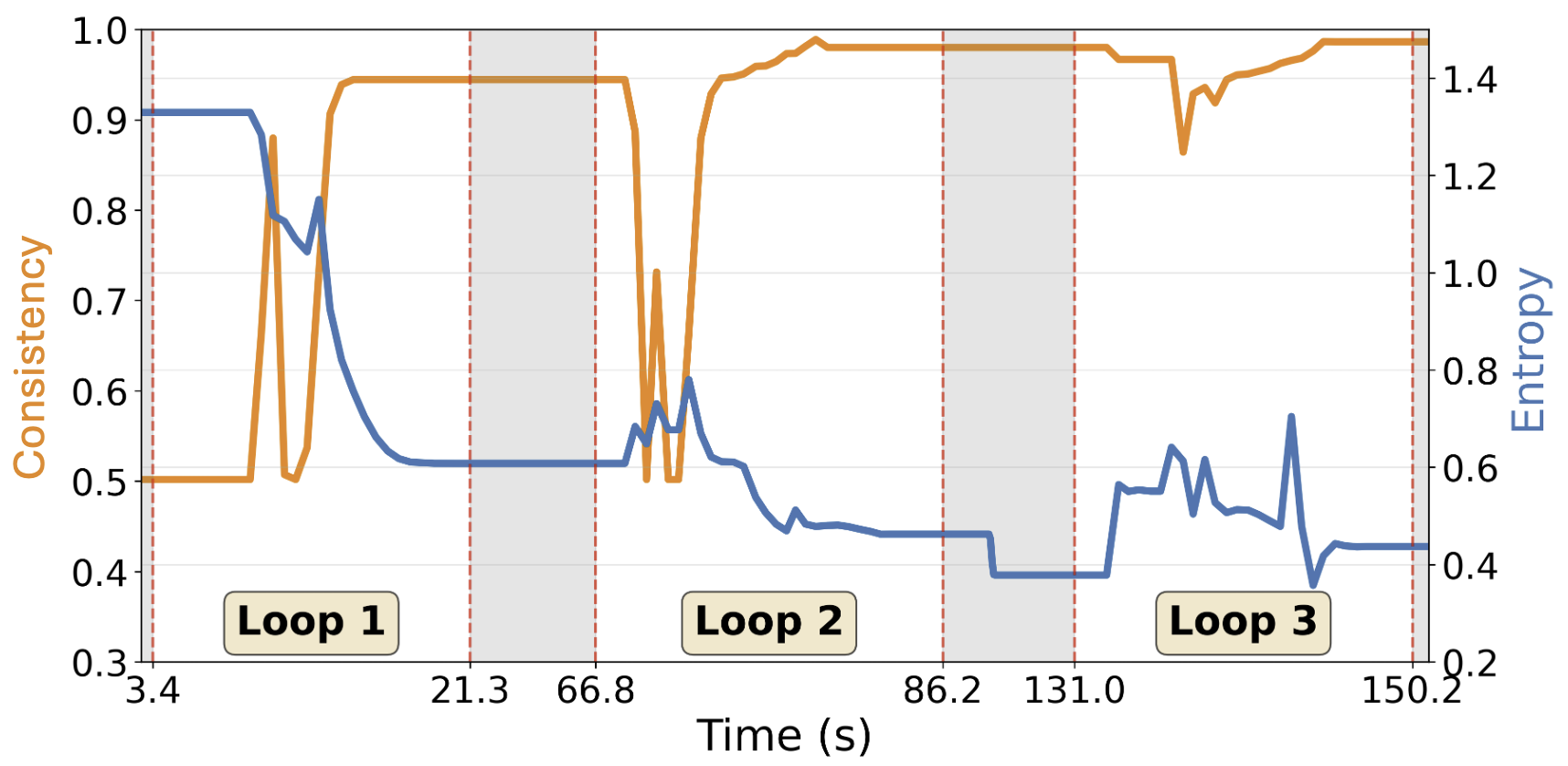}
        \caption{Box A.}
        \label{fig:semi_static_quant_a}
    \end{subfigure}
    
    \vspace{0.6em}
    
    \begin{subfigure}[b]{\columnwidth}
        \centering
        \includegraphics[width=\textwidth]{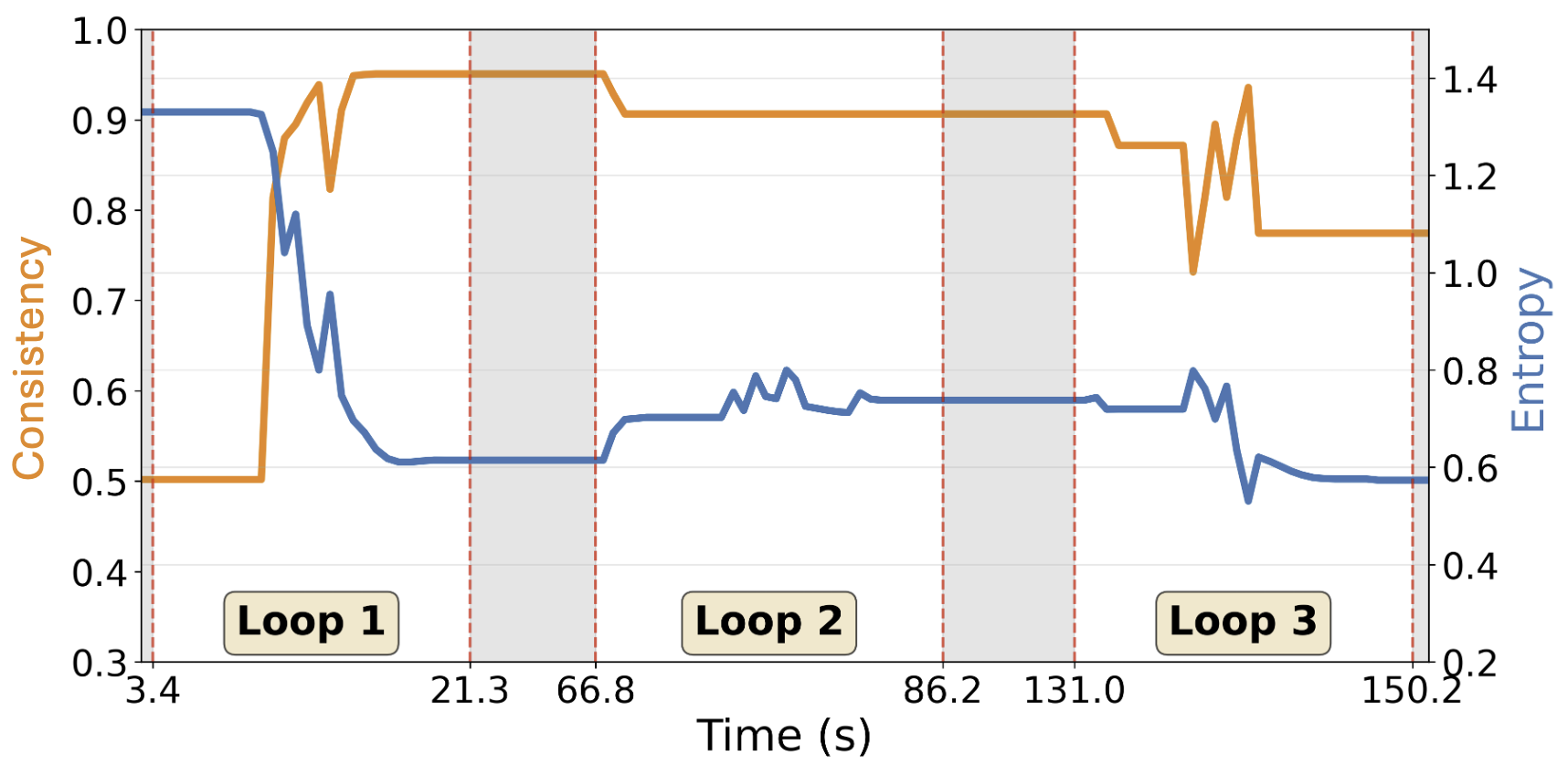}
        \caption{Box B.}
        \label{fig:semi_static_quant_b}
    \end{subfigure}
\caption{Consistency belief and task-dependent map entropy for two representative box regions in the controlled scene-change simulation shown in~\autoref{fig:semi_static_qual}. Vertical dashed lines mark loop boundaries, and shaded regions indicate intervals when the robot is away from the boxes and receives no observations of them. Each box changes primarily when its side of the corridor is displaced.}
\label{fig:semi_static_quant}
\end{figure}

\begin{figure*}[!t]
    \centering
    \includegraphics[width=0.95\linewidth]{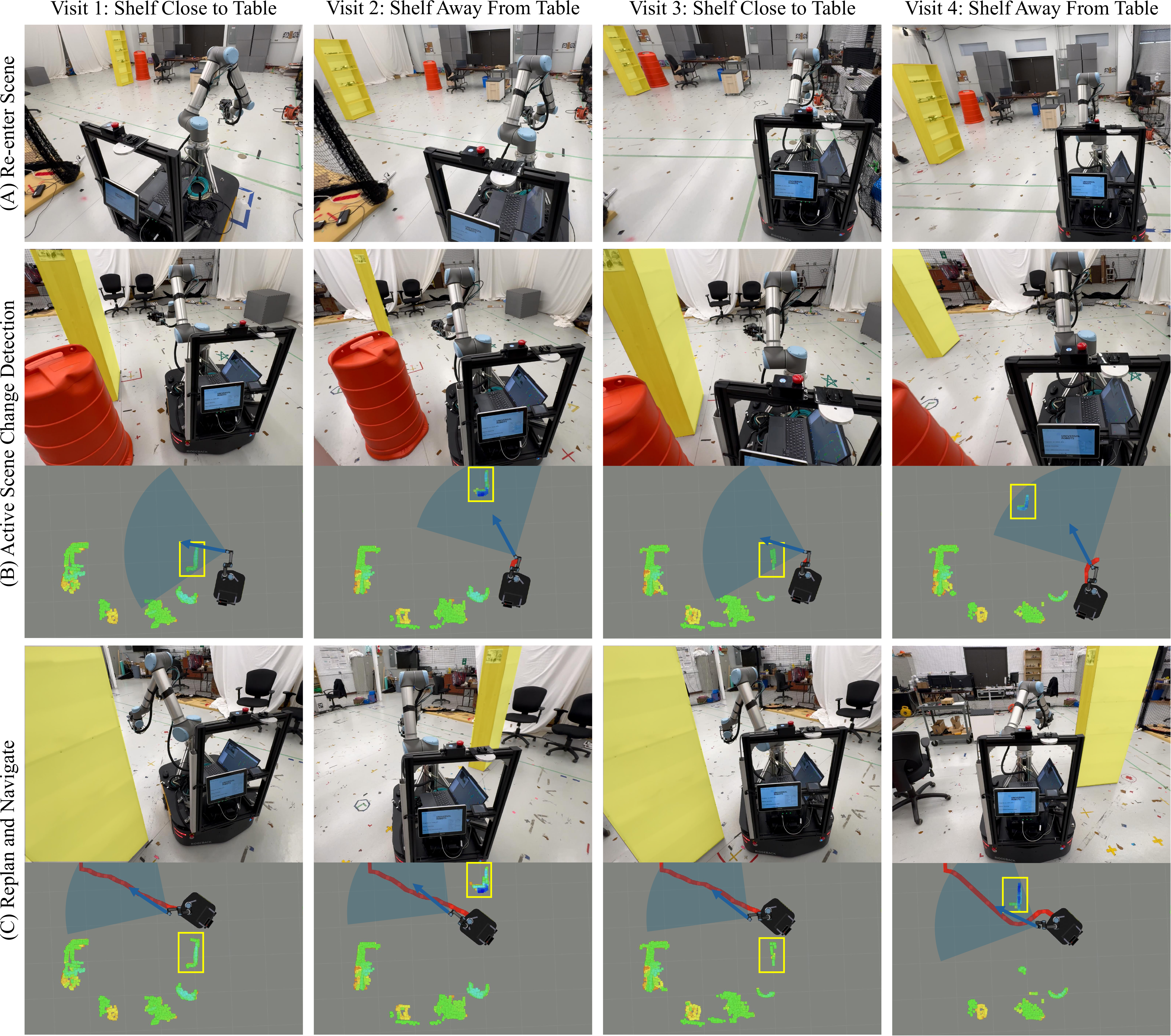}
\caption{Real-world long-term scene-change experiment over four visits. In each visit, the robot re-enters the room with the maintained object map, acquires new observations near the shelf, updates the map, and replans according to the current shelf position, switching between the outside and inner routes.\vspace{-0.3cm}}
\label{fig:office_scene_change}\vspace{-2mm}
\end{figure*}

The static benchmark demonstrates the benefit of observing missing geometry before a maneuver. We next use a controlled Isaac Sim experiment~\cite{NVIDIAIsaacSim} to examine how the object-aware uncertainty model responds when previously mapped objects move under a fixed navigation task. As shown in \autoref{fig:semi_static_qual}, the robot traverses a corridor formed by two rows of $1\mathrm{m}\times1\mathrm{m}$ boxes over three consecutive loops. Loop~1 starts from an empty map in the reference scene; Loops~2 and~3 start from the previous maintained map. Before Loop~2, boxes on the left are shifted, and before Loop~3, boxes on the right are shifted. The waypoint sequence and simulated environment otherwise remain unchanged, so the experiment focuses on object-level scene changes without the additional variability of the real-world trials. For this simulation, object and free-space confidence decay is turned off so that the observed changes come only from the object displacements.

\autoref{fig:semi_static_qual} shows the qualitative response of the policy. The red curves denote executed base trajectories, and the blue arrows denote sampled hand camera headings. In Loop~1, the hand camera largely follows the corridor direction. After the left-side boxes are shifted, the headings in Loop~2 bias toward the left. After the right-side boxes are shifted, the same behavior appears toward the right in Loop~3. Some views remain on the unchanged side because the robot continues the same navigation task, and localization errors and sensor noise can slightly reduce confidence in unchanged objects. Since the base trajectory remains similar across loops, the shift in viewing direction is not explained by the desired path alone. It indicates that inconsistency with the maintained map attracts additional sensing to the changed side, as intended by the object-aware NBV policy.

\autoref{fig:semi_static_quant} quantifies this behavior for two representative boxes. The dashed vertical lines mark loop boundaries, and the shaded regions mark intervals when the robot is away from the boxes and receives no direct observations. The orange curve shows the consistency belief $\mathbb{E}[v_i]$, and the blue curve shows average map entropy over a $1.5\mathrm{m}\times2.5\mathrm{m}$ horizontal rectangle around the box region, covering both the box and nearby free space before and after displacement. In Loop~1, the robot starts from an empty map, so the boxes are initialized only after first observation, with consistency belief $0.5$ and high associated map entropy. As observations accumulate, both boxes become more consistent with the maintained map. In Loop~2, Box~A is displaced: its consistency belief drops and entropy rises when observations first become inconsistent, then the object is re-localized and confidence recovers after more observations. Box~B remains comparatively stable. In Loop~3, the dominant change appears in Box~B after the right-side displacement; its consistency belief drops, while Box~A remains stable. The drop for Box~B does not trigger object reconstruction, whereas Box~A is re-localized after the earlier displacement. The average entropy around Box~B later decreases because moving the box farther from the robot exposes free space in front of it, lowering the regional mean even though the box consistency belief drops. The two objects respond in different loops, showing how map entropy evolves locally as changed objects are observed, consistent with \autoref{sec:method-posterior}.
\vspace{-1mm}

\subsection{Real-World Long-Term Scene Changes}

In the final experiment, we demonstrate the full closed-loop system under real-world long-term scene changes. \autoref{fig:office_scene_change} shows a four-visit sequence in the office scene (Scene~3 in \autoref{fig:real_world_scenes}), where a large shelf is moved closer to or farther from the table between traversals. This qualitative sequence examines how map updates guide sensing and replanning across visits, complementing the static-scene benchmark.

In all four visits, the robot is given the same waypoint sequence. Visit~1 starts from an empty map, and each later visit starts from the accumulated map of the previous visit. \autoref{fig:office_scene_change} pairs camera images with RViz views of the robot, maintained map, hand camera FoV, and planned base path; the shifted shelf is shaded in yellow. When the shelf is relocated, the maintained map becomes unreliable where the path may need to change. Guided by the proposed object- and task-aware NBV formulation, the robot first inspects free space upon re-entry and then returns to previously mapped objects near the shelf, as seen in the accompanying video. After new observations reveal inconsistency around the relocated shelf, map entropy rises there and attracts additional sensing to refine the map. The robot then replans according to the updated shelf position. In the figure, the hand camera turns toward the shelf region in all visits, while the planned path switches with the shelf position: when the shelf is close to the table in Visits~1 and~3, the robot drives around the shelf from the outside; when the shelf is farther away in Visits~2 and~4, it nudges through from the inner side. Together, the visits demonstrate the active perception and action loop targeted in this work. The robot carries a semi-static object-aware map across traversals, uses task-dependent map entropy to decide where the map should be checked or corrected, and uses the updated map to adapt navigation in a changing environment.

\section{Conclusions}\label{sec:conclusion}

We presented an object- and task-aware active perception framework for mobile manipulators operating in cluttered, semi-static environments. The main contribution of the paper is a probabilistic next-best-view formulation that combines maintained map uncertainty, object-level change awareness, and path-dependent task relevance to prioritize observations in regions most likely to affect safe navigation. This formulation is integrated with a hierarchical-task control framework so that the robot can realize informative hand camera motions while preserving the priority of the navigation task.

Real-world experiments demonstrate that the proposed framework reduces task-relevant map uncertainty, improves map consistency, and lowers navigation risk compared with passive or non-task-aware baselines. Overall, the results support the main premise of the paper: in semi-static environments, active perception should not be driven only by visibility or generic information gain, but by which uncertain and potentially changing regions matter most to safe progress along the planned path. By coupling semi-static map maintenance, task-aware viewpoint selection, and whole-body control in closed loop, the proposed system enables safer and more reliable navigation in changing environments across repeated visits.

\bibliographystyle{IEEEtran}
\bibliography{sections/references}

\end{document}